\documentclass{article}

\usepackage{microtype}
\usepackage{graphicx}
\usepackage{subfigure}
\usepackage{booktabs} 
\usepackage{adjustbox}
\usepackage{upgreek}
\usepackage{caption}
\usepackage{hyperref}
\usepackage{xcolor}
\usepackage{multirow}
\usepackage{enumitem}

\usepackage[accepted]{icml2024}

\usepackage{amsmath}
\usepackage{amssymb}
\usepackage{mathtools}
\usepackage{amsthm}
\usepackage{hyperref}

\usepackage[capitalize,noabbrev]{cleveref}

\theoremstyle{plain}
\newtheorem{theorem}{Theorem}[section]

\theoremstyle{definition}
\newtheorem{definition}[theorem]{Definition}

\theoremstyle{remark}

\usepackage[textsize=tiny]{todonotes}
\usepackage{tablefootnote}

\icmltitlerunning{Explaining Learned Reward Functions with Counterfactual Trajectories}

\begin{document}

\twocolumn[
\icmltitle{Explaining Learned Reward Functions with Counterfactual Trajectories}



\icmlsetsymbol{equal}{*}

\begin{icmlauthorlist}
\icmlauthor{Jan Wehner}{ci}
\icmlauthor{Frans Oliehoek}{tu}
\icmlauthor{Luciano Cavalcante Siebert}{tu}
\end{icmlauthorlist}

\icmlaffiliation{ci}{CISPA Helmholtz Center for Information Security}

\icmlaffiliation{tu}{Department of Intelligent Systems, Faculty of Electrical Engineering, Mathematics and Computer Science, Delft University of Technology, Delft, Netherlands}

\icmlcorrespondingauthor{Jan Wehner}{jan.wehner@cispa.de}

\icmlkeywords{Machine Learning, ICML}

\vskip 0.3in
]



\printAffiliationsAndNotice{}  

\begin{abstract}
Learning rewards from human behavior or feedback is a promising approach to aligning AI systems with human values but fails to consistently extract correct reward functions. Interpretability tools could enable users to understand and evaluate possible flaws in learned reward functions.
We propose Counterfactual Trajectory Explanations (CTEs) to interpret reward functions in Reinforcement Learning by contrasting an original and a counterfactual trajectory and the rewards they each receive.
We derive six quality criteria for CTEs and propose a novel Monte-Carlo-based algorithm for generating CTEs that optimizes these quality criteria.
To evaluate how informative the generated explanations are to a proxy-human model, we train it to predict rewards from CTEs.
CTEs are demonstrably informative for the proxy-human model, increasing the similarity between its predictions and the reward function on unseen trajectories. Further, it learns to accurately judge differences in rewards between trajectories and generalizes to out-of-distribution examples. 
Although CTEs do not lead to a perfect prediction of the reward, our method, and more generally the adaptation of XAI methods, are presented as a fruitful approach for interpreting learned reward functions and thus enabling users to evaluate them.
\end{abstract}

\section{Introduction}
As Reinforcement Learning (RL) models grow in their capabilities and adoption in real-world applications \cite{yu2021healthcare, kiran2022auronomous, afsar2022recommender}, we must ensure that they are safe and aligned with human values. A core difficulty of achieving trustworthy and controllable AI \cite{cavalcante2023meaningful, russell2019human} is to accurately capture human intentions and preferences in the reward function on which the RL agent is trained since the reward function will shape the agent's objectives and behaviour.
For many tasks, it is hard to manually specify a reward function that accurately represents the intentions, preferences, or values of designers, users or society at large \cite{pan2022effects, amodei2016concrete}. Reward Learning is a set of techniques that circumvents this problem by instead learning the reward function from data. 
For example, Preference-based RL \cite{christiano2017pbrl} derives a reward function from preference judgments queried from a human and has recently been applied to control the behaviour of Large Language Models \cite{bai2022trainingshort}. Similarly, Inverse RL \cite{ng2000algorithms}, which is commonly used in autonomous driving and robotics, aims to retrieve the reward function of an expert from the demonstrations they generate. 
Reward learning is a promising approach for aligning the reward functions of AI systems with the intentions of humans \cite{russell2019human, leike2018scalable}. It has significant advantages over behavioral cloning, which learns a policy by using supervised learning on observation-action pairs since reward functions are considered the most succinct, robust, and transferable definition of a task \cite{abbeel2004apprenticeship}. However, these techniques suffer from a multitude of theoretical \cite{armstrong2018occam, skalse2022misspecification} and practical problems \cite{casper2023openshort} that make them unable to reliably learn human values which are diverse \cite{lera2022towards}, dynamic \cite{vandepoel2024change} and context-dependent \cite{liscio2022values}.

We aim to develop interpretability tools that help humans to understand learned reward functions so that they can detect misalignments with their own values.
This is in line with the ``Transparent Value Alignment" framework in which Sanneman and Shah \cite{sanneman2023transparent} suggest leveraging techniques from eXplainable AI (XAI) to provide explanations about the reward function. The process of explaining reward functions can be useful for both the \textit{understanding} and \textit{explaining} phases of the XAI pipeline~\cite{dwivedi2023explainable}, by enabling both developers and users to inspect reward functions. This is a relevant task for the XAI community, as it contributes to the goal of enabling human users to understand, appropriately trust, and produce more explainable models ~\cite{dwivedi2023explainable,sanneman2023transparent}. However, there have been few attempts to interpret reward functions and only Michaud et al. \cite{michaud2020understanding} attempt this for deep, learned reward functions. Our work makes a novel connection between XAI and reward learning by providing, to the best of our knowledge, the first principled application of counterfactual explanations to reward functions.

\begin{figure}
    \centering
    \includegraphics[width=\linewidth]{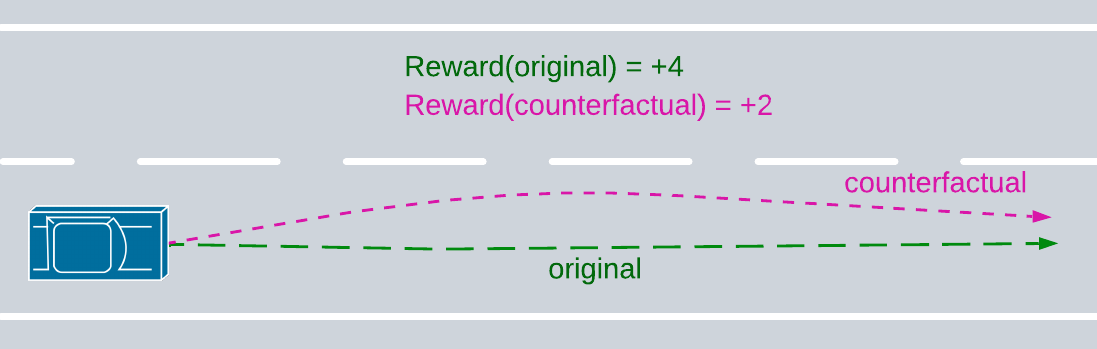}
    \caption{A car has originally taken a straight line and received a reward of $+4$ from the reward function. By providing a counterfactual that receives a lower reward of $+2$ the user can make hypotheses about how the reward function assigns rewards.}
    \label{fig:car_example}
\end{figure}

Counterfactual explanations are a popular XAI tool that has not yet, to the best of our knowledge, been applied to explain reward functions. It helps humans to understand the predictions of ML models by posing hypothetical ``what-if" scenarios. Humans commonly use counterfactuals for decision-making, learning from past experiences, and emotional regulation\cite{byrne2016counterfactual, kahneman1986norm, roese2014might}. Thus users can intuitively reason about and learn from counterfactual explanations, which makes this an effective and user-friendly mode of explanation \cite{mittelstadt2019explaining, wachter2018counterfactual, mandel2011causal}. 

\textbf{We propose Counterfactual Trajectory Explanations (CTEs) that serve as informative explanations about deep reward functions.} CTEs can be employed in a sequential decision-making setting by contrasting an original with a counterfactual partial trajectory along with the rewards assigned to them. This enables the user to draw inferences about what behaviours cause the reward function to assign high or low rewards. For instance, consider the domain of autonomous driving illustrated in Figure \ref{fig:car_example}. While a given driving trajectory by itself might not provide much insight, adding a counterfactual trajectory along with its reward allows a user to hypothesise that the reward function negatively rewards the driving agent for swerving and getting close to the other lane.

In order to generate CTEs we identify and adapt six quality criteria for counterfactual explanations from XAI and psychology and introduce two algorithms for generating CTEs that optimise for these quality criteria. To evaluate how effective the generated CTEs are we introduce a novel measure of informativeness in which a proxy-human model learns from the provided explanations. Implementation details, ablations and further experiments can be found in the technical appendix.  \footnote{The full code for the project is available at: \href{https://github.com/janweh/Counterfactual-Trajectory-Explanations-for-Learned-Reward-Functions}{https://github.com/janweh/Counterfactual-Trajectory-Explanations-for-Learned-Reward-Functions}}

\section{Counterfactual Trajectory Explanations (CTEs)}
\label{sec:preliminaries}
This study focuses on adapting counterfactual explanations to interpret a learned reward function. Counterfactual explanations alter the inputs to a given system, which causes a change in the outputs \cite{wachter2018counterfactual}. When explaining reward functions the inputs could either be single states or (partial) trajectories. Correspondingly, the outputs to be targeted can either be seen as rewards assigned to single states or as the average reward assigned to the states in a (partial) trajectory.
If we would only alter individual states, multi-step plans could be overlooked and infeasible counterfactuals that cannot occur through any sequence of actions might be created. By generating trajectories and showing their average rewards we can provide the user with insights about which multi-step behaviours are incentivized by the reward function, while also guaranteeing that counterfactuals are feasible. While it would be possible to generate multiple counterfactuals per original, we only show the user one counterfactual to be able to cover more original trajectories. 

We operate in Markov Decision Processes consisting of states $S$, actions $A$, transition probabilities $P$ and a reward function $R$. Further, we denote a learned reward function as $R_\theta: S \times A \Rightarrow \mathbb{R}$,  a policy trained for $R_\theta$ as $\pi_\theta$, full trajectories generated by a full play-through of the environment as $\tau$ and partial trajectories as $t \subseteq \tau$. Counterfactual Trajectory Explanations (CTEs) can now be defined as:

\begin{definition}
\label{def:cte}
CTEs $\{(t_{org}, \overline{r}_{org}),(t_{cf}, \overline{r}_{cf})\}$ consist of an original and counterfactual partial trajectory and their average rewards assigned by a reward function $R_\theta$. Both start in the state $s_n$ but then follow a different sequence of actions resulting in different average rewards.

\end{definition}

The difference in rewards can be causally explained by the difference in actions. If the agent had chosen actions $(a_{cf_n}, ...,a_{cf_{k}})$ instead of $(a_{org_n}, ...,a_{org_{m}})$ resulting in $t_{cf}$ instead of $t_{org}$ the reward function $R_\theta$ would have assigned an average reward $\overline{r}_{cf}$ instead of $\overline{r}_{org}$.\footnote{Examples of CTEs in the Emergency Environment \cite{peschl2022moral} can be found in: \href{https://drive.google.com/drive/folders/1JMjwQM24BbDwL8vRnG3pST5hlvpzRfZM?usp=sharing}{https://drive.google.com/drive/folders\newline /1JMjwQM24BbDwL8vRnG3pST5hlvpzRfZM?usp=sharing}}

We propose a method to address the following problem:
Given a learned reward function $R_\theta$, a policy $\pi_\theta$ trained on $R_\theta$ and a full original trajectory $\tau_{org}$ generated by $\pi_\theta$, the task is to select a part of that trajectory $t_{org} \subseteq \tau_{org}$ and generate a counterfactual $t_{cf}$ to it that starts in the same state $s_n$ so that the resulting CTE is informative for an explainee to understand $R_\theta$.

\section{Method}
This Section presents the method used to generate CTEs. First, quality criteria that measure the quality of an explanation are derived from the literature and combined into a scalar quality value. Then two algorithms are introduced which generate CTEs by optimising for the quality value.

\label{sec:method}
\subsection{Determining the quality of CTEs}
\label{subsec:quality-criteria}
Counterfactual explanations are usually generated by optimising them for a loss function that determines how good a counterfactual is \cite{artelt2019computation}. This loss function combines multiple aspects, which we call ``quality criteria".

\subsubsection{Quality Criteria}
\label{subsubsec:quality_criteria}
By reviewing XAI literature we were able to identify 9 quality criteria that are used for counterfactual explanations. These criteria are designed to make counterfactuals more informative to a human. Out of these \textsf{\small Causality}, \textsf{\small Resource} and \textsf{\small Actionability} \cite{keaneif, verma2018programmatically, gajcin2022counterfactual} are automatically achieved by our methods. We are left with six quality criteria to optimise for which we adapt to judge the quality of CTEs.

\textbf{1. \textsf{\small Validity}:} Counterfactuals should lead to the desired difference in the output of the model \cite{verma2018programmatically, gajcin2022counterfactual}. This difference in outputs makes it possible to causally reason about the changes in the inputs. We maximise \textsf{\small Validity} as $\left | R_\theta(t_{org}) - R_\theta(t_{cf}) \right |$.

\textbf{2. \textsf{\small Proximity}:} The counterfactual should be similar to the original \cite{keaneif, MILLER20191, gajcin2022counterfactual}. Thus we minimize a measure based on the Modified Hausdorff distance \cite{dubuisson1994hausdorff} that finds the closest match between the state-actions pairs in the two trajectories. The distance of state-action pairs is calculated as a weighted sum of the Manhattan distance of the player positions, whether the same action was taken and the edit distance between non-player objects in the environment.

\textbf{3. \textsf{\small Diversity}:}
Explanations should cover the space of possible variables as well as possible \cite{Huang.2019b, frost2022explaining}. Consequently, each new CTE should establish novel information rather than repeating previously shown CTEs. 
Thus we maximize \textsf{\small Diversity} of a new CTE compared to previous CTEs. This is calculated as the sum of the average difference between the new length of the trajectory and previous lengths, the average difference in the new starting time in the environment and previous starting times, and the fraction of previous trajectories that are of the same counterfactual direction.
Counterfactual direction can be upward or downward comparisons \cite{roese1994functional} when the reward of the counterfactual is higher or lower than the original's reward.

\textbf{4. \textsf{\small State importance}:} Counterfactual explanations should focus on important states that have a significant impact on the trajectory outcome \cite{frost2022explaining}. We aim to start counterfactual trajectories in critical states, where the policy strongly favors some actions over others. We maximize the importance of a starting state which is calculated as the policies negative entropy $-\sum_{a \in A} \pi(a|s_0) \log \pi(a|s_0)$ \cite{frost2022explaining, huang2018establishing}.

\textbf{5. \textsf{\small Realisticness}:}
The constellation of variables in a counterfactual should be likely to happen \cite{keaneif, gajcin2022counterfactual, verma2018programmatically}. In our setting, we want counterfactual trajectories that are likely to be generated by a policy trained on the given reward function. Such a trajectory would likely score high on the reward function. Thus we maximize: $\overline{R_\theta(t_{cf})} - \overline{R_\theta(t_{org})}$.

\textbf{6. \textsf{\small Sparsity}:}
Counterfactuals should only change a few features compared to the original to make it cognitively easier for a human to process the differences \cite{keaneif, verma2018programmatically, gajcin2022counterfactual, MILLER20191}. Instead of meticulously restricting the number of features that differ between states we lighten the cognitive load by incentivizing CTEs to be short by minimizing: $len(t_{org}) + len(t_{cf})$.

\subsubsection{Combining quality criteria into a scalar quality value}
After measuring the six quality criteria, we scalarise them into one \textit{quality value} $\rho$ to be assigned to a CTE. This is done by normalising the criteria and combining them into a weighted sum. Criteria are normalised to $[0,1]$ by iteratively generating new CTEs with random weights and adapting the minimum and maximum value the criteria take on.

The weights $\upomega$ assigned to the quality criteria correspond to their relative importance. However, this opens the question of how one should weigh the different quality criteria to generate the most informative explanations for a certain user. 
To find the optimal set of weights we suggest a \textit{calibration phase} in which $N$ different sets of weights $\upomega=\{\omega_{Validity_j}, ..., \omega_{Sparsity_j}\}^{N}_{j=1}$ are uniformly sampled  $\omega_i \sim U(0,1)$ and used to create CTEs. The CTE's informativeness is tested and the set of weights that produces the most informative CTEs to a specific user are chosen for further use.

\subsection{Generation algorithms for CTEs}
\label{subsec:generation}
In order to generate CTEs we propose two algorithms that optimise for the aforementioned quality value (see Section \ref{subsec:quality-criteria})  along with a random baseline algorithm. 

\textbf{Algorithm 1 - Monte Carlo-based Trajectory Optimization (\textsf{\small MCTO}):} \\
\textsf{\small MCTO} adapts Monte Carlo Tree Search (MCTS) to the task of generating CTEs. MCTS is a heuristic search algorithm that has been applied to RL by modelling the problem as a game tree, where states and actions are nodes and branches \cite{silver2016mastering, vodopivec2017monte}.
It uses random sampling and simulations to balance exploration and exploitation in estimating the Q-values of states and actions.

In contrast to MCTS, \textsf{\small MCTO} operates on partial trajectories instead of states, optimises for quality values instead of rewards from the environment, adds a termination action which ends the trajectory and applies domain-specific heuristics. Pseudocode \ref{alg:mcto} showcases the algorithm.

\begin{algorithm}[tb]
   \caption{Monte Carlo Trajectory Optimization}
   \label{alg:mcto}
\begin{algorithmic}
    \STATE {\bfseries Input:} full trajectory $\tau_{org}$, environment $env$, actions $A$
    \STATE $candidates =$ [] \hfill \% store candidate CTEs
    \FOR{$s_n$ in $\tau_{org}$}
        \STATE $Q$ = [] \hfill \% Q-values of trajectories
        \STATE $t_{cf}$ = [$s_n$]
        \REPEAT
            \FOR{$i$ to $n_{iterations}$}
                \STATE $t_{cf}^s$ $\leftarrow$ SELECTION($t_{cf}$)
                \STATE $t_{cf}^e$ $\leftarrow$ EXPANSION($t_{cf}^s$)
                \STATE $\rho$ $\leftarrow$ SIMULATION($t_{cf}^e$)
                \STATE $Q$ $\leftarrow$ BACK-PROPAGATION($Q$, $\rho$)
            \ENDFOR
            \STATE $a^*=argmax_{a \in A}(Q(t_{cf}, a))$
            \STATE $s_n$ $\leftarrow$ $env$.step($s_n$, $a^*$)
            \STATE APPEND($t_{cf}, (s_n, a^*)$)
        \UNTIL{$s_n$ is $terminal$}
        \STATE $t_{org}$ = SUBSET($\tau_{org}, s_n, |t_{cf}|$) \hfill \% Subtrajectory from \\ \hfill \% $s_n$ with same lengths as $t_{cf}$
        \STATE APPEND($candidates, (t_{org}, t_{cf})$)
    \ENDFOR
    \STATE {\bfseries Return:} $argmax_{c \in candidates}\rho(c)$
    
\end{algorithmic}
\end{algorithm}

In \textsf{\small MCTO} nodes represent partial trajectories $t$, branches are actions $a$ and child nodes result from parents by following the action in the connecting branch.
Leaf nodes are terminated trajectories which can occur from entering a terminal state in the environments or by selecting an additional terminal action that is always available. \textsf{\small MCTO} optimises for the quality value $\rho$ of a CTE, which is being measured at the leaf nodes. 
A CTE is derived by taking the partial trajectory in the leaf node as the counterfactual $t_{cf}$ and the subtrajectory of $\tau_{org}$ from starting state $s_n$ with the same length as $t_{cf}$ as the original $t_{org}$.

Each state $s_n \in \tau_{org}$ in the original trajectory is used as a potential starting point of the CTE by setting it as the root of the tree and running \textsf{\small MCTO}. Out of these, the CTE with the highest quality value is chosen.
For a given state we choose the next action by repeating these four steps for a set number of times ($n_{iterations}$) before choosing the action $a^*$ with the highest Q-value:
\begin{enumerate}
    \item \uppercase{Selection}: A node in the tree, which still has unexplored branches is chosen. The choice is made according to the \textit{Upper Confidence Bounds for Trees} algorithm based on the estimated Q-value of the branches and the number of times the nodes and branches have already been visited.
    \item \uppercase{Expansion}: After selecting a node, we choose a branch and create the resulting child node.
    \item \uppercase{Simulation}: One full playout is completed by sampling actions uniformly until the environment terminates the trajectory or the terminating action is chosen. At each step, the terminal action is chosen with a probability of $p_{MCTO}(end)$.  The resulting CTE's quality value $\rho$ is evaluated according to the quality criteria.
    \item \uppercase{Back-propagation}: $\rho$ is back-propagated up the tree to adjust the Q-values of previous nodes $t$: $Q(t) = \frac{1}{N(t)}(\rho - Q(t))$.
\end{enumerate}

As an efficiency-increasing heuristic, we prune off branches of actions that have a likelihood $\pi_\theta(a|s) \le threshold_a$ to be chosen by the policy. Furthermore, we choose not to employ a discount factor ($\gamma = 1$) when back-propagating $\rho$, since this would incentivize shorter CTEs while this is already done by the \textsf{\small Sparsity} criterion. Ablations showed that other heuristics such as choosing actions in the simulation based on the policy $\pi_\theta$ or basing the decisions for expansion on an early estimate of the $\rho$ did not improve performance.

\textbf{Algorithm 2 - Deviate and Continue (DaC):}\\
The \textit{Deviate and Continue} (\textsf{\small DaC}) algorithm creates a counterfactual trajectory $t_{cf}$ by deviating from the original trajectory $\tau_{org}$ before continuing by choosing actions according to policy $\pi_\theta$. Starting in a state $s_n \in \tau_{org}$, the deviation is performed by sampling an action from the policy $\pi_\theta$ that leads to a different state than in the original trajectory. After $n_{deviations}$ such deviations $t_{cf}$ is continued by following $\pi_\theta$. During the continuation, there is a $p_{DaC}(end)$ chance per step of ending both $t_{org}$ and $t_{cf}$. This process is repeated for every state $s_n \in \tau_{org}$ and the resulting CTE with the highest quality value is chosen.

\textbf{Baseline Algorithm - Random}
As a weak baseline, we compare our algorithms to randomly generated CTEs. A start state $s_n$ of the counterfactual is uniformly chosen from the original trajectory $\tau_{org}$. From there actions are uniformly sampled, while the trajectories have a  $p_{Random}(end)$ chance of being ended in each timestep.

\section{Evaluation}
\label{sec:exp}
\begin{figure}
    \centering
    \includegraphics[width=\linewidth]{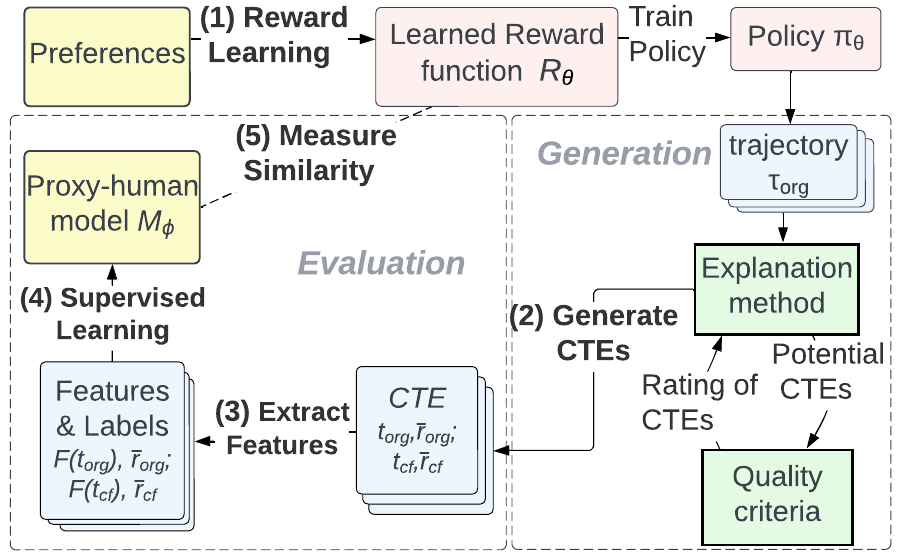}
    \caption{Schematic that describes how rewards are learned (1), explanations are generated (2) and evaluated (3,4\&5).}
    \label{fig:full-schematic}
\end{figure}

This Section details the experimental approach we take to evaluate the informativeness of CTEs. 
We want to automatically measure how well an explainee can understand a reward function from explanations, while similar works perform user studies or do not offer quantitative evaluations. Since previous methods for interpreting reward functions are not applicable to our evaluation setup we can only compare our proposed methods with a baseline and criteria with each other. 
Our evaluation approach includes learning a reward function, generating CTEs about it and measuring how informative the CTEs are for a proxy-human model (see Figure \ref{fig:full-schematic}).

\subsection{Generating reward functions and CTEs}
To learn a reward function (1) we first generate expert demonstrations. A policy $\pi^*$ is trained on a ground-truth reward $R^*$ via Proximal Policy Optimization (PPO) \cite{schulman2017proximal}. This policy is used to generate $1000$ expert trajectories $\uptau_{exp} = \{\tau_{exp_k}\}^{1000}_{k=1}$. 
Secondly, we use Adversarial IRL \cite{fu2017learning} which derives a robust reward function $R_\theta$ and policy $\pi_\theta$ from the demonstrations by posing the IRL problem as a two-player adversarial game between a reward function and a policy optimizer. 

We use the Emergency environment \cite{peschl2022moral}, a Gridworld environment that represents a burning building where a player needs to rescue humans and reduce the fire. The environment 7 humans that need to be rescued, a fire extinguisher which can lessen the fire and obstacles which block the agent from walking through. In each timestep, the player can walk or interact in one of the four directions.
This environment is computationally cheap and simple to investigate. However, it is still interesting to study since the random initialisations require the reward function to generalise while taking into account multiple sources of reward.

To make CTEs about $R_\theta$ (2) we first generate a set of full trajectories $\uptau_{org}=\{\tau_{org_k}\}^{1000}_{k=1}$ using the policy $\pi_\theta$.
Lastly, we use the algorithms described in Section \ref{subsec:generation} to optimise for the quality criteria in Section \ref{subsec:quality-criteria} to produce one CTE per full trajectory $CTEs = \{t_{org_k}, t_{cf_k}\}_k^{1000}$.
We conducted a grid search of hyperparameters for each of the generation algorithms. Based on that we choose $p_{MCTO}(end)=0.35$, $threshold_a = 0.003$ and $n_{iterations} = 10$ for \textsf{\small MCTO}, $p_{DaC}(end)=0.55$ and $n_{deviaitons}=3$ for \textsf{\small DaC} and $p_{Random}(end)=0.15$ for \textsf{\small Random}.

\subsection{Evaluating the informativeness of CTEs}
\label{subsec:evaluation}
We argue that informative explanations allow the explainee to better understand the learned reward function, which we formalize as the explainee's ability to assign similar average rewards to unseen trajectories as the reward function.

To evaluate the informativeness of CTEs, we employ a Neural Network (NN) as a proxy-human model to learn from the explanations and to predict the average reward assigned by $R_\theta$ for a trajectory. While humans learn differently from data than an NN, this evaluation setup still gives us important insights into the functioning and effectiveness of CTEs. 

Notably, this measure only serves to evaluate the generation method and would not be used when showing CTEs to humans. It allows us to test whether extracting generalisable knowledge about the reward function from the provided CTE is possible by measuring how well the
proxy-human model can predict unseen CTEs. Furthermore, it allows us to compare different algorithms and quality criteria by measuring and contrasting the informativeness of CTEs they generate.

The evaluation procedure consists of three steps, as presented in Figure \ref{fig:full-schematic}: (3) features and labels are extracted from the CTEs to form a dataset to train on, (4) a proxy-human model is trained to predict the rewards of trajectories from these features, and, lastly, (5) the similarity between the predictions of the proxy-human model and the rewards assigned by $R_\theta$ is measured to indicate how informative the CTEs were to the model.

\textbf{Extracting features and labels (3)}\\
We extract 46 handcrafted features $F(t)= \{f_0, ..., f_{45}\}$ about the partial trajectories. These features represent concepts that the reward function might consider in its decision-making, for example of the form ``time spent using item X'' or ``average distance from object Y''. We opted against methods for automatic feature \cite{8938371} extraction to avoid introducing more moving parts in the evaluation.
The average reward for the states in a partial trajectory serves as the label for the proxy-human model $\overline{r}=\frac{1}{|t|}\Sigma_{s \in t} R_\theta(s)$. By averaging the reward we avoid biasing the learning to the length of partial trajectories.

\textbf{Learning a proxy-human model (4)}\\
A proxy-human regression model $M_\phi$ is trained to predict the average reward $\overline{r}$ given to the partial trajectory $t$ by $R_\theta$ from the extracted features $F(t)$. Humans learn from counterfactual explanations in a contrastive manner by looking at the difference in outputs to causally reason about the effect of the inputs \cite{MILLER20191} but also learn from the individual data points. Since we aim to make $M_\phi$ learn in a similar way to a human we train $M_\phi$ on two tasks.
In the \textit{single task}, it is trained to separately predict the average reward for the original and the counterfactual. Giving rewards to unseen trajectories shows how similar the judgements of $M_\phi$ and $R_\theta$ are for trajectories. The loss on one CTE for this task is the sum:
$L_{single}(t_{org}, t_{cf})=(M_\phi(t_{org}) - R_\theta(t_{org}))^2 + (M_\phi(t_{cf}) - R_\theta(t_{cf}))^2$.\\
In the \textit{contrastive task}, $M_\phi$ is trained to predict the difference between the average original and counterfactual reward. By doing this we train $M_\phi$ to reason about how the difference in inputs causes the outputs instead of only learning from data points independently: $L_{contrastive}(t_{org}, t_{cf})=[(M_\phi(t_{org}) - (M_\phi(t_{cf})) - (R_\theta(t_{org}) - R_\theta(t_{cf})]^2$.

$M_\phi$ is defined as a 4-layer NN that receives the features extracted from both the original and the counterfactual as a concatenated input and is trained in a multi-task fashion on single and contrastive tasks. The body of the NN is shared between both tasks and feeds into two separate last layers that perform the two tasks separately.
The losses of both tasks are used separately to update their respective last layer and are added into a weighted sum to update the shared body of the network.

We train the NN on 800 samples with the Adam optimiser and weight decay and results are averaged over  30 random initialisations. We perform hyperparameter tuning using 5-fold cross-validation for the learning rate, regularisation values, number of training epochs and dimensionality of hidden layers.

\textbf{Measuring similarity to the reward function (5)}\\
To measure how similar the proxy-human model's predictions are to the reward function's outputs we measure the Pearson Correlation between them on unseen CTEs. Reward functions are invariant under multiplication of positive numbers and addition \cite{ng1999invariance}. This is well captured by the Pearson Correlation because it is insensitive to constant additions or multiplications. To ensure a fair comparison between different settings we test how well a model trained on CTEs from one setting generalises to a combined test set that contains CTEs from all settings.

\section{Experiments}
\label{sec:results}
This Section describes the results of three experiments that test the overall informativeness of CTEs, compare the generation algorithms and evaluate the quality criteria.

\subsection{Experiment 1: Informativeness of Explanations for proxy-human model}
\label{subsec:exp-1}

\textbf{Experimental Setup: }
We want to determine the success of our methods in generating informative explanations for a proxy-human model $M_\phi$, while also comparing the generation algorithms on the downstream task.
As described in Section \ref{subsec:evaluation} each generation algorithm produced 800 CTEs on which we trained 10 $M_\phi$s each, before testing the Pearson Correlation between their predictions and the average rewards on a combined test set of 600 CTEs. We use the weights from Table \ref{tab:opt_weight} for the quality criteria.

\begin{figure}
    \centering
    \includegraphics[width=\linewidth]{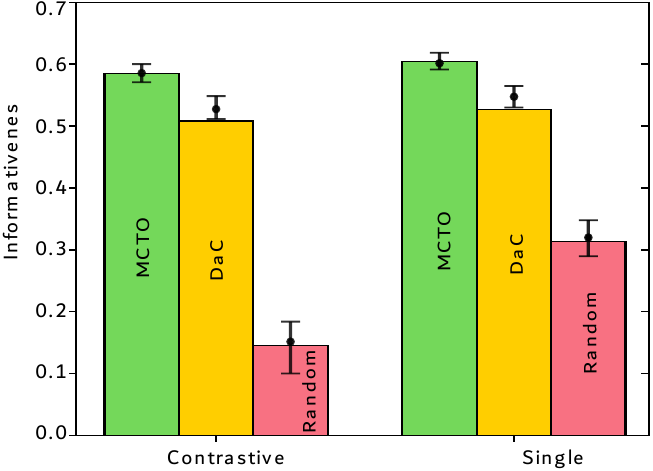}
    \caption{The average informativeness of CTEs generated by \textsf{\small MCTO}, \textsf{\small DaC} and \textsf{\small Random} for a NN trained for \textit{single} and \textit{contrastive} predictions, along with median, upper and lower quartile.}
    \label{fig:methods_informativeness_combined}
\end{figure}

\textbf{Results: }
Figure \ref{fig:methods_informativeness_combined} shows that $M_\phi$s trained on CTEs from \textsf{\small MCTO} achieved on average higher correlation values. $M_\phi$s trained on \textsf{\small DaC}'s CTEs were significantly ($p<0.001$) worse, while the models trained on randomly generated CTEs achieved a much lower correlation on both tasks. 

\subsection{Experiment 2: Quality of Generation Algorithms}
\label{subsec:exp-2}
\textbf{Experimental Setup: }
This experiment tests how good the generation algorithms are at optimising for the quality value. Each generation algorithm produced 1000 CTEs and their quality value $\rho$ was measured. To make this test independent of the weights for quality criteria, each CTE is optimised for a different uniformly sampled set of weights: $\upomega=\{\omega_{Validity_j}, ..., \omega_{sparsity_j}\}^{1000}_{j=1}$, where  $\omega_i \sim U(0,1)$. Furthermore, the efficiency of algorithms (seconds/generated CTE) and the length and starting time of CTEs were recorded.

\textbf{Results: }
From Table \ref{tab:qc-methods} we see that \textsf{\small MCTO} achieved a higher average quality value than \textsf{\small DaC}, which again outperformed the random baseline (differences are significant with $p<1e^{-7}$).
However, the higher performance came at a computational cost, since \textsf{\small MCTO} was slower, while \textsf{\small Random} was very efficient.
On average the trajectories of \textsf{\small Random} were the longest and those of \textsf{\small MCTO} the shortest.
Lastly, both \textsf{\small MCTO} and \textsf{\small DaC} tended to choose starting times earlier in the environment (20.96 and 20.45 out of 75 timesteps).

\begin{table}[]
    \centering
\begin{tabular}{|l|ccc|}
\hline
                 &   \textsf{\small MCTO} &   \textsf{\small DaC} &   \textsf{\small Random} \\
\hline
 Avg quality value $\rho$ $\uparrow$&   \textbf{1.44} &  1.32 &     1.1  \\
 Std quality value $\rho$ &0.47&0.49&0.37\\
 \hline
 Efficiency (s/CTE) $\downarrow$ \tablefootnote{Efficiency differs depending on the hardware used.}&  14.86 &  5.46 &     \textbf{0.04}\\
 Length (\# steps)&   2.76 &  4.96 &     7.41 \\
 Starting Points (\# first step)&  20.96 & 20.45 &    42.58 \\
\hline
\end{tabular}
    \caption{Shows the average quality value $\rho$ and its variance achieved by \textsf{\small MCTO}, \textsf{\small DaC} and \textsf{\small Random}, along with the efficiency of generating CTEs, the length of the CTEs and at what step in the environment they started.}
    \label{tab:qc-methods}
\end{table}

\subsection{Experiment 3: Informativeness of quality criteria}
\label{subsec:exp-3}
\textbf{Experimental Setup: }
Finally, we wanted to determine the influence of a quality criterion on informativeness. For this, we analyzed the Spearman correlation between the weight assigned to the criterion during the generation of a set of CTEs and the informativeness of this set of CTEs.
Simultaneously we carried out the calibration phase to determine the set of weights which leads to the most informative CTEs for an explainee and generation algorithm.

Thirty sets of weights $\upomega$ were each used to generate one set of 1000 CTEs with \textsf{\small MCTO}. 800 CTEs were used to train 10 $M_\phi$s as described in Section \ref{subsec:evaluation}. The performances of the resulting 30 sets of $M_\phi$s were evaluated on a test set that combines the remaining 200 samples from each of the 30 sets of CTEs. This indicates the informativeness of the CTEs they were trained on.
By measuring the Spearman correlation between the weights assigned to a criterion and the informativeness of the resulting CTEs for $M_\phi$, we can infer the importance of that criterion for making CTEs informative. Furthermore, we record the set of weights which leads to the most informative CTEs for each generation algorithm except \textsf{\small Random} which is independent of weights.

\begin{table}[]
    \centering
    \setlength{\tabcolsep}{3pt}
    \begin{tabular}{|c c c c c c|}
    \hline
        Validity & Proximity & Diversity & State Importance & Realisticness & Sparsity \\
        
        0.982& 0.98 & 0.576 & 0.528 & 0.303 & 0.851\\
        \hline
    \end{tabular}
    \caption{Most informative set of weights for \textsf{\small MCTO} and \textsf{\small DaC}.}
    \label{tab:opt_weight}
\end{table}

\begin{figure}
    \centering
    \includegraphics[width=.9\linewidth]{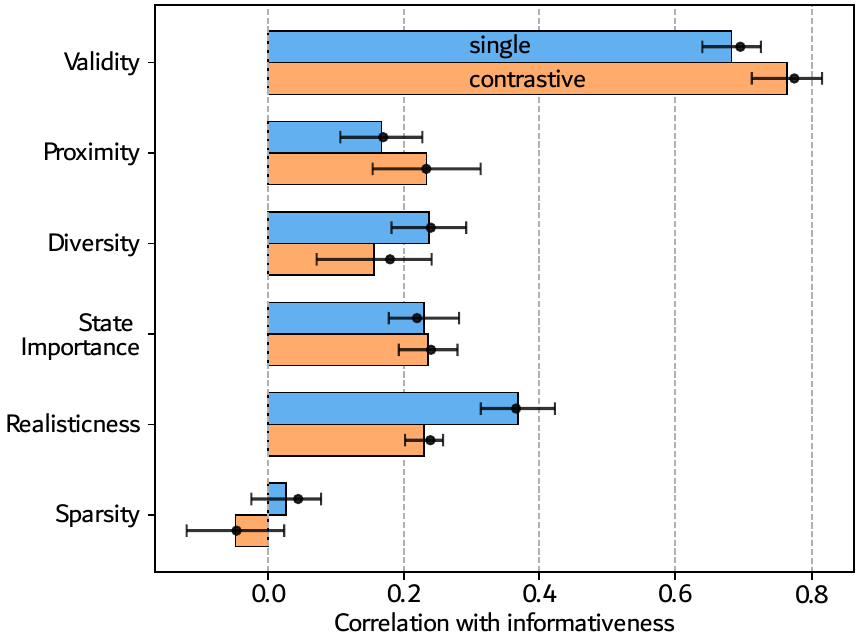}
    \caption{Spearman correlation between weights for the quality criteria and the informativeness of the resulting CTEs for $M_\phi$ for the contrastive and single task. Averaged over 10 models along with the median and upper and lower quartile.}
    \label{fig:weights-step}
\end{figure}

\textbf{Results: }
Figure \ref{fig:weights-step} shows that for both contrastive and single learning, the weights of \textsf{\small Validity} ($\omega_{Validity}$) correlated the strongest with the informativeness for $M_\phi$. This is followed by $\omega_{Realisticness}$, $\omega_{Proximity}$, $\omega_{Diversity}$ and $\omega_{State Importance}$ which all show a moderate correlation with the informativeness, while $\omega_{Sparsity}$ was barely or even negatively correlated with informativeness. While there are differences between the importance of criteria for the two tasks, they end up with similar results.

Furthermore, we find that the same set of weights leads to the most informative CTEs for both \textsf{\small MCTO} and \textsf{\small DaC}. It assigns very high weights to \textsf{\small Validity} and \textsf{\small Proximity}, while \textsf{\small Realisticness} is weighted low. Contrary to Figure \ref{fig:weights-step} \textsf{\small Sparsity} is highly weighted.

\subsection{Discussion}
\textbf{CTEs are informative for the proxy-human model.} Experiment 1 shows that an NN-based model trained on CTEs is much better than random guessing at predicting rewards or judging the difference in rewards between unseen CTEs. It also shows a capability to generalise to out-of-distribution examples when predicting CTEs generated by other algorithms. This indicates that CTEs enable an explainee to learn some aspects of the reward function which hold generally across different distributions of trajectories.

However, the fact that the correlations of $M_\phi$'s predictions with the true labels are $\leq 0.60$ clearly shows that there are aspects of the reward function, which $M_\phi$ did not pick up on. This could be explained by a lack of training samples, a loss of information during the feature extraction or insufficient coverage of different situations in the environment. Furthermore, the studied reward function is noisy, often outputting different rewards for apparently similar situations and is thus hard to understand.

\textsf{\small MCTO} generated the most informative CTEs, while the CTEs from \textsf{\small Random} were less informative.

Similarly, we find that \textbf{\textsf{\small MCTO} is the most effective generation algorithm for optimising the quality value}, while \textsf{\small DaC} outperforms \textsf{\small Random}. The fact that the algorithms which achieved higher quality values in Experiment 2 also produced more informative CTEs in Experiment 1 indicates that optimising well for the quality value is generally useful for making more informative CTEs.
Table \ref{tab:qc-methods} shows a trade-off, between the performance and efficiency of the generation algorithms, which likely appears because a more exhaustive search finds higher-scoring CTEs.
Furthermore, \textsf{\small MCTO} and \textsf{\small DaC} selected CTEs with earlier starting times. This is because the environment had higher fluctuations in rewards early on, which benefits \textsf{\small \textsf{\small Validity}} and \textsf{\small State importance}. This shows that they are able to select CTEs in more interesting parts of the environment.
They also tend to choose shorter trajectories, which score higher on \textsf{\small Sparsity}.

Among the criteria \textbf{\textsf{\small Validity} is the most important criterion for generating informative CTEs} as shown in Experiment 3.
High weights for \textsf{\small Validity} lead to higher differences in rewards and lead to a larger range of labels for contrastive predictions. Possibly, an NN can learn more information from these larger differences and is thus better informed by CTEs that are high in \textsf{\small Validity}.
\textsf{\small Proximity}, \textsf{\small Realisticness}, \textsf{\small Diversity} and \textsf{\small State importance} are also beneficial for having the proxy-human model learn from CTEs, but we are less certain about why they are beneficial. Although prioritising \textsf{\small Sparsity} does not 
correlate with informativeness, the most informative set of weights does give it a high weight. However, this high weight might be a fluke since we only tried 30 sets of weights. In any case, we should not conclude that humans would not benefit from sparse explanations. While NNs can easily compute gradients over many different features simultaneously, humans can only draw inferences about a few features at once \cite{miller1956magical}. This clarifies that the prioritisation of quality criteria will likely differ for a human.

The fact that the two tasks largely agreed on the importance of quality criteria indicates that they complement each other. This might be because the two tasks are similar and thus benefit from developing similar representations in the shared body of the network. Furthermore, because the same set of weights out of 30 options led to the most informative CTEs when using \textsf{\small MCTO} and \textsf{\small DaC} we can speculate that the relative importance of quality criteria for an explained is similar, independent of the generation algorithm used.

\textbf{Limitations:} Since we do not measure the informativeness of CTEs for a human user, our experiments do not prove that CTEs are informative for humans or show how important the criteria would be to a user. Furthermore, we only conduct experiments on a single learned reward function in a single environment, making it unclear how our findings will generalise to other settings. The method might especially struggle with large and complex environments where it is difficult to achieve high coverage of the environment with CTEs. Further, depends on the ability to reset the environment to previous states, which is not given in some environments. Lastly, our evaluation measure depends on hand-crafted features which limits its applicability.

\section{Related Work}
This Section covers previous work on the interpretability of reward functions and counterfactual explanations for AI. 
\label{sec:related_work}

\subsection{Interpretability of Learned Reward Functions}
Reward functions can be made intrinsically more interpretable by learning them as decision trees \cite{bewley2022interpretable, kalra2022differentiable, srinivasan2020interpretable} or in logical domains \cite{kasenberg2017interpretable, munzer2015inverse}. 
Attempts have been made to make deep reward functions more interpretable by simplifying them through equivalence transformation \cite{jenner2022preprocessing}  or by imitating a Neural Network with a decision tree \cite{russell2019explaining}.
However, such interpretable representations can negatively impact the performance of the method.

To avoid this drawback, we interpret learned reward functions via post-hoc explanations. Post-hoc methods are applied after the model has been trained to explain the model's decision-making process.
Lindsey and Shah \cite{sanneman2021explaining, sanneman2022empirical} test the effectiveness and required cognitive workload of simple explanation techniques about linear reward functions. While their work requires linear reward functions our method is applicable to any representation of a reward function.

The closest work to ours comes from Michaud et al. \cite{michaud2020understanding} who apply gradient salience and occlusion maps to identify flaws in a learned reward function and employ handcrafted counterfactual inputs to validate their findings.
Our work focuses on counterfactuals and automatically generates them to be of high quality.

\subsection{Counterfactual Explanations}
Despite a large body of work on generating counterfactual explanations about ML models in supervised learning problems \cite{verma2020counterfactual, artelt2019computation, guidotti2022counterfactual, stepin2021survey} and their relation to human psychology \cite{keaneif, byrne2019counterfactuals}, this approach has only recently been adapted to explain RL policies. Counterfactuals consist of a change in certain input variables which cause a change in outputs \cite{wachter2018counterfactual}. In the RL setting, counterfactual explanations can be changes in Features, Goals, Objectives, Events, or Expectations that cause the agent to change its pursued Actions, Plans, or Policies \cite{gajcin2022counterfactual}. 
This can improve users' understanding of out-of-distribution behaviour \cite{frost2022explaining}, provide them with more informative demonstrations \cite{9982062} or showcase how an agent's environmental beliefs influence its planning \cite{NEURIPS2021_926ec030}.
Instead of explaining a policy $\pi$ this paper presents the first principled attempt to use them to use counterfactuals to explain a reward function $R$.

\section{Conclusion}
\label{sec:conclusion}
While reward learning presents a promising approach for aligning AI systems with human values, there is a lack of methods to interpret the resulting reward functions. To address this we formulate the notion of Counterfactual Trajectory Explanations (CTEs) and propose algorithms to generate them. Our results show that CTEs are informative for an explainee, but do not lead to a perfect understanding of the reward function. Further, they validate our \textsf{\small MCTO} algorithm to be effective at generating CTEs and imply that the difference in outcomes between an original and counterfactual trajectory is especially important to achieve informative explanations.
This research demonstrates that it is fruitful to apply techniques from XAI to interpret learned reward functions.

Future work should carry out a user study to test the informativeness of CTEs for humans. 
Furthermore, the method should be evaluated in more complex environments and on a range of reward functions produced by different reward learning algorithms.
Ultimately, we hope that CTEs will be used in practice to allow users to understand the misalignments between their values and a reward function, thus enabling them to improve the reward function with new demonstrations or feedback.

\section{Acknowledgements}
The project on which this report is based was funded by the Federal Ministry of Education and Research under the funding code 16KIS2012. Further, it was partially supported by TAILOR, a project funded by EU Horizon 2020 research and innovation programme under GA No 952215.


\bibliography{article.bib}
\bibliographystyle{icml2024}

\newpage
\appendix
\onecolumn

\section{Environment}
\label{app:environment}
\begin{figure}[h!]
    \centering
    \includegraphics[height=5cm]{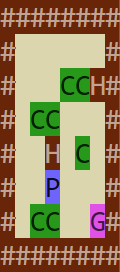}
    \caption{A random initialisation of the Emergency environment by Peschl et al. \yrcite{peschl2022moral}. Shows the player \textit{P} in blue, the humans \textit{C} in green, the obstacles \textit{H} in brown, the fire-extinguisher \textit{G} in pink and the borders of the environment \textit{\#} in brown.}
    \label{fig:environment}
\end{figure}

For our experiments, we employ the \textit{Emergency} environment developed by Peschl et al. \yrcite{peschl2022moral} which is a Gridworld environment that represents a burning building where a player needs to rescue humans and reduce the fire.
The environment contains a player, 7 humans that need to be rescued, a fire extinguisher which can lessen the fire and obstacles which block the agent from walking through (see Figure \ref{fig:environment}). Humans can be rescued by standing next to them and interacting with their cells. The fire extinguisher is placed in the bottom right and used by standing on its cell. At each step, the agent can either move in one of the four directions, interact with an adjacent cell or stand still. In each run, the starting position of the player, humans and obstacles are randomly initialised and the environment is run for 75 steps. A ground truth reward function $R^*$ assigns a reward of $+10$ per human saved and $+1$ per step the fire extinguisher is used.

We choose this environment since it is simple, but still requires generalisation and provides multiple sources of reward. The fact that it is a deterministic, small Gridworld environment makes it computationally cheaper to train PPO and AIRL and thus allows for faster iterations in the development cycle. Since the starting positions of the agent, humans and obstacles are randomised the policy and reward function are required to have different initialisations. Lastly, providing multiple sources of rewards means that the learned reward function needs to capture multiple aspects of the environment, which makes it more interesting to provide explanations for it.

The environment is available at: \href{https://github.com/mlpeschl/moral_rl}{https://github.com/mlpeschl/moral\_rl}

\section{Additional implementation details}
\label{app:quality_criteria}
\subsection{Implementation of quality criteria}
\label{app:impl_qual}
In this appendix, we lay out the implementation of the quality criteria we use. For the motivation of the criteria see Section \ref{subsubsec:quality_criteria}.

\subsubsection{\textsf{\small Validity}}
We calculate the difference in rewards between $t_{org}$ and $t_{cf}$ as the absolute difference in average rewards: \textsf{\small Validity} as $\left | R_\theta(t_{org}) - R_\theta(t_{cf}) \right |$, where $R_\theta(t)=\frac{1}{|t|}\Sigma_{s \in t}R_\theta(s)$.  The average reward attained is used to not bias the \textsf{\small Validity} with regards to the differences in length between trajectories.

\subsubsection{Proximity}
We interpret \textsf{\small Proximity} as the similarity between trajectories, which is measured using the  Modified Hausdorff distance (MHD) \cite{dubuisson1994hausdorff}. MHD is a technique to calculate the similarity between two sets of points and has been used in previous works in IRL to evaluate whether generated trajectories are similar to each other. For two trajectories $A$ and $B$, it calculates how large the distance of each point in $A$ is to its closest point in $B$ and vice versa and then takes the higher of those two values as the distance between the trajectories:

$MHD(d(A,B),d(B,A)) = max(d(A,B),d(B,A))$, where
$d(A,B) = \frac{1}{N_a}\sum_{a \in A} min_{b \in B} \text{dist}(a,b)$

The distance between state action pairs $dist(a,b)$ takes into three factors:
\begin{itemize}
    \item Manhattan distance between the player coordinates in a and b to measure how far the player is apart in these states $dist_{player}(a,b) = |x_a - x_b| + |y_a - y_b|$, where $x$ and $y$ denote the players x- and y-coordinates in the environment in that state.
    \item Whether the action taken was the same $dist_{action}(a,b) = \begin{cases} 
    0 & \text{if } a_{action} = b_{action} \\
    1 & \text{otherwise}
    \end{cases}$.
    \item  The edit distance between humans $dist_{humans}(a,b) = ||a_{humans} - b_{humans}||$, where $||.||$ indicates the edit distance operation and $b_{humans}$ denotes the x- and y-coordinates of all humans in the environment.
\end{itemize}

These are then added into a weighted sum: $dist(a,b) = 1.5 \cdot dist_{player}(a,b) + 0.5 \cdot dist_{action}(a,b) + dist_{humans}(a,b)$. 
The weight between the factors was determined by trying out different weights and subjectively judging how the weighted sum matched with the researcher's judgment about the similarity between trajectories. Furthermore, the distance metric is not environment-agnostic, since we draw on our knowledge about the specific environment to determine which factors should go into the distance measurement of states.

\subsubsection{Diversity}
We calculate \textsf{\small Diversity} as the sum of 4 aspects of trajectories: their length, starting time, starting state and whether they are an upward or downward counterfactual. \textsf{\small Diversity} is calculated for each potential new CTE in relation to previous CTEs that have already been shown to the user. So for the first CTE \textsf{\small Diversity}$=0$ and for the 5th CTE, it is measured with regards to the 4 previously generated CTEs.

To measure how similar the length of the counterfactual is to the lengths of previous counterfactuals we take the three previous CTEs which are closest in length (smallest difference) and calculate their average difference in length to the new CTE. In the same way, we compare the starting time of a CTE to the starting times of previous CTEs by comparing the 3 closest previous CTEs and calculating their average difference in starting times.

Further, counterfactuals can be divided into upward and downward counterfactuals, depending on whether the counterfactual leads to a better (up) or worse (down) outcome than the original \cite{roese1994functional}. To add more diversity we add a reward based on how often the type of counterfactual has already been shown previously. If the currently considered CTE is an upward CTE we reward it if few of the previous CTEs were upwards $\frac{down}{down+up}$ and if it is downwards we do the opposite $\frac{up}{down+up}$.

Finally, these criteria are summed together to form one \textsf{\small Diversity} criterion.

\subsubsection{State importance}
States can be considered critical if the variance in the Q-function for the actions or the entropy in the policies distribution over actions could serve as heuristics for interesting states.  Amir and Amir \cite{amir2018highlights}  calculate the importance of a state as the difference in Q-values between the best and the worst action: $I(s) = max_aQ^\pi_{(s,a)} - min_aQ^\pi_{(s,a)}$, while 
Huang et al. \cite{huang2018establishing} take the difference between the best and average action $I(s) = max_aQ^\pi_{(s,a)} - \frac{1}{|A|}\sum_{a \in A}Q^\pi_{(s,a)}$. Both Forst et al \cite{frost2022explaining} and Huang et al. \cite{huang2018establishing} use the negative entropy in the probability distribution over actions in that state \textsf{\small State Importance}$(s) = -\sum_{a \in A} \pi(a|s) \log \pi(a|s)$. 
Since our method does not produce Q-values, we rely on the negative entropy version to measure how critical the starting state of a CTE is and select CTEs that score high on this criterion.

\subsubsection{Realisticness}
We interpret \textsf{\small Realisticness} as meaning that a trajectory is likely to be generated by a policy trained on the reward function that is being explained. Intuitively a trajectory is more likely if it achieves a high reward on the reward function. To make this unbiased to the length of the trajectory we take the average reward per step. Further, to not bias the criteria to parts of the environment where high rewards are easily achieved we use the reward achieved by the original trajectory as a comparison:  \textsf{\small Realisticness}$=\overline{R_\theta(t_{cf})} - \overline{R_\theta(t_{org})}$.

\subsubsection{Sparsity:} We interpret this as meaning that the counterfactual and original trajectories should be shorter so that humans do not have to compare many behaviours and states. Thus we sum up the length of the original and counterfactual: \textsf{\small Sparsity}$(t_{org}, t_{cf}) = len(t_{org}) + len(t_{cf})$.\\

\subsection{Normalisation of quality criteria}
\label{app:normalising}
After we have measured the quality criteria of a CTE, we want to combine them into a scalar by taking the weighted sum. Before doing so we normalise each criterion to a range of $[0,1]$. This requires finding a maximum ($norm_{max}^c$) and minimum ($norm_{min}^c$) possible value that each criterion $c$ can take on. We do this approximately in an adaptive fashion, by 
\begin{enumerate}
    \item Starting with $norm_{max}^c=1$ and $norm_{min}^c=0$.
    \item Using these normalisation values DaC and MCTO generate 20 CTEs for a random set of weights.
    \item Finding the minimum $norm^{c\prime}_{min}=min$ and maximum $norm_{max}^{c\prime}=1$ values for $c$ in the CTEs.
    \item Updating the normalisation parameters if more extreme values were found \\
$norm_{max}^c= \max(norm_{max}^c, norm_{max}^{c\prime})$ and $norm_{min}^c= \max(norm_{min}^c, norm_{min}^{c\prime})$.
    \item If the normalisation values were adapted for any $c$: Repeat from Step 2.
\end{enumerate}
When performing this procedure for our experiments, we also manually altered hyperparameters and settings of the generation algorithms between repeats to account for the fact that some settings can lead to more extreme values. 

\subsection{Pseudocode for Deviate and Continue}
\label{app:pseudo-dac}
Pseudocode \ref{alg:dac} shows how \textsf{\small DaC} is implemented.

\begin{algorithm}[tb]
   \caption{Deciate and Continue}
   \label{alg:dac}
\begin{algorithmic}
    \STATE {\bfseries Input:} full trajectory $\tau_{org}$, environment $env$, actions $A$, policy $\pi_\theta$, $n_{deviations}$
    \STATE $candidates =$ [] \hfill \% store candidate CTEs
    \FOR{$(s_{org}^n,a^n_{org})$ in $\tau_{org}$}
        \STATE $s_{cf} = s_{org}^n$
        \STATE $t_{cf}$ = [$s_{cf}$]
        \FOR{$i$ in $[0,n_{deviations})$}
        \STATE \hfill \% DEVIATE from the original
            \REPEAT 
               \STATE $a_{cf}= \pi_\theta(a \in A | s_{cf})$ \hfill \% Resample $a_{cf}$ until it leads to a different state\\
                \STATE $s_{cf}$ = $env$.step($s_{cf}, a_{cf}$)
            \UNTIL{$s_{cf} \neq s_{org}^{n+i+1}$} 
            \STATE APPEND($t_{cf}$, ($s_{cf}$, $a_{cf}$))
        \ENDFOR
        \STATE $a_{cf}$ = $\{ a_{end} \text{ with } p_{DaC}(end)$ $|$ $\pi_\theta(a|s) \text{ otherwise }\}$ \hfill \% CONTINUE using the policy $\pi_\theta$
        \WHILE{$a_{cf} \neq a_{end}$} 
            \STATE $s_{cf}$ = $env$.step($s_{cf}$, $a_{cf}$)
            \STATE APPEND($t_{cf}$, ($s_{cf}$, $a_{cf}$))
            \STATE $a_{cf}$ = $\{ a_{end} \text{ with } p_{DaC}(end)$ $|$ $\pi_\theta(a|s) \text{ otherwise }\}$ \hfill \% End trajectory with likelihood $p_{DaC}(end)$
        \ENDWHILE
        \STATE $t_{org}$ = SUBSET($\tau_{org}, s_{org}^n, |t_{cf}|$) \hfill \% Subtrajectory from $s_{org}^n$ with same lengths as $t_{cf}$
        \STATE APPEND($candidates, (t_{org}, t_{cf})$)
    \ENDFOR
    \STATE {\bfseries Return:} $argmax_{c \in candidates}\rho(c)$ \hfill \% Return the highest rated candidate
    
\end{algorithmic}
\end{algorithm}

\subsection{Implementation of Extracted Features and Labels}
\label{app:extracted_features}
After the CTEs were generated we extracted features from them, which are then used to train the proxy-human model. For both the original and counterfactual partial trajectory we extract 46 handcrafted features $F(t)= \{f_0, ..., f_{45}\}$ and the average rewards $\overline{r}$. Here $t=\{(s_n,a_n), ..., (s_m, a_n)\}$ denotes a partial trajectory with state action pairs with length $|t|=m-n$. Furthermore,  in a state $s$ we denote $humans(s)$ as the number of humans that are still in danger,  $H(s)$ as a list of the x-y-coordinates of all unsaved humans and $p_x(s)$ and $p_y(s)$ as the x- and y-coordinate of the player.

We only extract features on the level of partial trajectories, not for single states. One can expect that a significant amount of detail gets lost when only considering features on the trajectory level since some aspects which factor into the rewards for single states get lost.
However, we want the proxy-human model to make predictions about partial trajectories and not only single states so that it can evaluate multi-step behaviour. Since trajectories can have different lengths we cannot simply feed it all single-state features, but need to accumulate them into a feature vector of the same dimensionality.
The features are normalised across the set of CTEs to have a mean of 0 and a standard deviation of 1.

\textbf{List of Features:}\\
Features about humans:
\begin{enumerate}
    \item Humans saved: How many humans did the agent save during the partial trajectory? $humans(s_m) - humans(s_n)$.
    \item Unsaved humans: On average over the steps in the trajectory how many humans were not saved yet? $\frac{1}{|t|}\Sigma^m_{i=n} humans(s_i)$.
    \item Final number of unsaved humans: At the end of the partial trajectory, how many humans are not saved yet? $humans(s_m)$.
    \item Number of humans: In how many of the steps in the trajectory were there $n \in \{0, ..., 7\}$ humans still in danger? For each $n$: $\Sigma_{i=n}^m \begin{cases} 
    1 & \text{if } humans(s_i)=n \\
    0 & \text{otherwise}
    \end{cases}$.
    \item Average Distance between humans: How far are the humans apart in Euclidean distance averaged over all pairs of humans? $\frac{1}{humans(s)^2} \Sigma_{(h_x,h_y) \in H(s)}\Sigma_{(h_x^\prime,h_y^\prime) \in H(s)} |h_x-h_x^\prime|+|h_y-h_y^\prime|$.
\end{enumerate}

Features about the fire extinguisher, which is located at the x-y-coordinates $(6,6)$:
\begin{enumerate}[resume]
    \item Standing on fire extinguisher: For how many steps did the agent use the fire extinguisher? $\Sigma_{i=n}^m \begin{cases} 
    1 & \text{if } p_x(s_i)=6 \wedge p_y(s_i)=6 \\
    0 & \text{otherwise}
    \end{cases}$
    \item Distance to fire-extinguisher: On average what is the Euclidean distance between the agent and fire-extinguisher? $\frac{1}{|t|}\Sigma^m_{i=n} |p_x(s_i) - 6| + |p_y(s_i) - 6|$
\end{enumerate}

Features about the actions of the agent:
\begin{enumerate}[resume]
    \item Could have saved: Sums over the trajectory in how many states the agent stood next to a human and could have saved them, but failed to do so. This means in the state $s_i$ the Euclidian distance of the player to a human was exactly 1 and in the next state $s_{i+1}$ the amount of unsaved humans is still the same as in $s_i$.
    \item Moved towards the closest human: How often did the agent move towards the closest human or save a human? This is true if the Euclidian distance to the closest human $s_i$ is larger than in the following state $s_{i+1}$. There is a special case when the player is standing on the human since it can only interact with the human while standing \textit{next} to them. Since it would take 2 actions to save this human (step aside and interact with them) we calculate this as a distance of 2 to this human. Furthermore, it is true if the number of humans in $s_i$ is larger than in $s_{i+1}$.
    \item Type of action: How often did the agent take a type of action walking, interacting or standing still? 
\end{enumerate}

Features about the distance of the agent to humans, where:
\begin{enumerate}[resume]
    \item  Average Distance to humans: On average how close was the agent to the closest human? $\frac{1}{|t|}\Sigma^m_{i=n} min_{(h_x,h_y) \in H(s_i)} |p_x(s_i) - h_x| + |p_y(s_i) - h_y|$.
    \item Distances to humans: In how many steps in the trajectory was the agent $d \in \{0, ..., 10\}$ steps away from the closest human.
    \item Direction of human: How often was the closest human right, left, down or up from the agent? For each of the directions, it counts in how many states the closest human was in that direction. If the closest human is to the top-left it counts towards both up and left.
\end{enumerate}

Features about the position of the agent?
\begin{enumerate}[resume]
    \item Position of the agent: Average x and y position of the agent. $\frac{1}{|t|}\Sigma^m_{i=n} p_x(s_i)$ (equivalent for y).
    \item Steps next to the wall: For how many steps was the agent next to the wall? Counts for how many of the states the agent was positioned next to the wall as the number of states $s \in t$ which fulfil: $p_x(s)=1 \lor p_x(s)=6 \lor p_y(s)=1 \lor p_y(s)=6$. 
    \item Steps in the middle: Counts for how many states $s \in t$ the agent stood in the middle, thus fulfilling: $(p_x(s),p_y(s)) \in \{(3,3),(3,4),(4,3),(4,4)\}$.
    \item Steps between middle and wall: For how many states $s \in t$ was the agent not in the middle or next to the wall?
    \item Time spent in quadrant: For how many steps was the agent in the top-left ($p_x\leq 3 \wedge p_y\leq 3$), top-right ($p_x> 3 \wedge p_y\leq 3$), bottom-left ($p_x\leq 3 \wedge p_y> 3$) or bottom-right quadrant ($p_x> 3 \wedge p_y> 3$)? 
 \end{enumerate}
 
Other:
\begin{enumerate}[resume]
    \item Length: Number of steps in the partial trajectory $|t|$.
\end{enumerate}

\subsection{Hyperparameters for AIRL \& PPO}
\label{app:hyperparameters}
Table \ref{tab:ppo} shows the hyperparameters that were used to train PPO on the ground-truth reward function $R^*$ and generated the demonstrations which $R_\theta$ was learned from.

Table \ref{tab:airl} displays the hyperparameters for Adversarial Inverse Reinforcement Learning that is used to learn a reward function $R_\theta$ from the demonstrations. Furthermore, it also learns a policy $\pi_\theta$ alongside the reward function, which is used to generate original trajectories and is sometimes referred to during the generation of CTEs.

\begin{table}[h]
    \begin{minipage}{0.5\textwidth}
        \centering
        \begin{tabular}{|c|c|}
        \hline
        Hyperparameter & Value \\
        \hline
        Environment Steps & 6e6 \\
        Learning Rate & 1e-4 \\
        Batch Size & 12 \\
        epochs & 5 \\
        $\gamma$ & 0.999 \\
        Entropy Regularization & 0.1 \\
        $\epsilon$-clip& 0.1 \\
        \hline
        \end{tabular}
        \caption{Hyperparameters of PPO used to generate the demonstrations}
        \label{tab:ppo}
    \end{minipage}
    \begin{minipage}{0.5\textwidth}
        \centering
        \begin{tabular}{|c|c|}
        \hline
         Hyperparameter&  Value\\
         \hline
         Environment Steps & 3e6\\
         Learning Rate Discriminator & 5e-4 \\
         Learning Rate PPO & 1e-5 \\
        Batch Size Discriminator & 12 \\
        Batch Size PPO & 12 \\
        PPO epochs & 5 \\
        $\gamma$ & 0.999 \\
        $\epsilon$-clip& 0.1 \\
        \hline
        \end{tabular}
        \caption{Hyperparameters for Adversarial Inverse Reinforcement Learning}
        \label{tab:airl}
    \end{minipage}
\end{table}

\section{Influence and Trade-offs in quality criteria}
\label{app:trade-offs}
This section explores what trade-offs and synergies exist between the quality criteria, how these influence the choice of CTEs and how the different generation algorithms perform on the quality criteria.\\

\subsection{Trade-offs and influence of quality criteria}
\label{app:trade-offs-exp}
Maximising the quality value $\rho$, which combines six quality criteria, is a multi-objective optimization problem, where the different aspects of the objective are combined into a decision by taking the normalised, weighted sum of criteria (see Appendix \ref{app:normalising}). Thus we want to analyse the trade-offs and synergies between criteria. To study this we consider the correlations between quality criteria of the candidate CTEs, not only over the CTEs that are finally chosen.

\textbf{Experimental Setup: }
\textsf{\small DaC} is employed to generate 1 CTE for each of the 1000 sets of weights $\upomega_{random}$. 
For each run of \textsf{\small DaC} we measure the quality criteria and calculate their Pearson correlation between each quality criterion. We then average the resulting correlations over the 1000 runs.

\textbf{Results:} There are significant negative correlations between \textsf{\small \textsf{\small Validity}} and \textsf{\small Proximity}, \textsf{\small Diversity} and \textsf{\small Sparsity}. \textsf{\small Diversity} is further negatively correlated with \textsf{\small State Importance} and \textsf{\small Sparsity}. Lastly, there is a positive correlation between \textsf{\small \textsf{\small Validity}} and \textsf{\small State Importance}.

\begin{figure}
    \centering
    \includegraphics[width=0.5\linewidth]{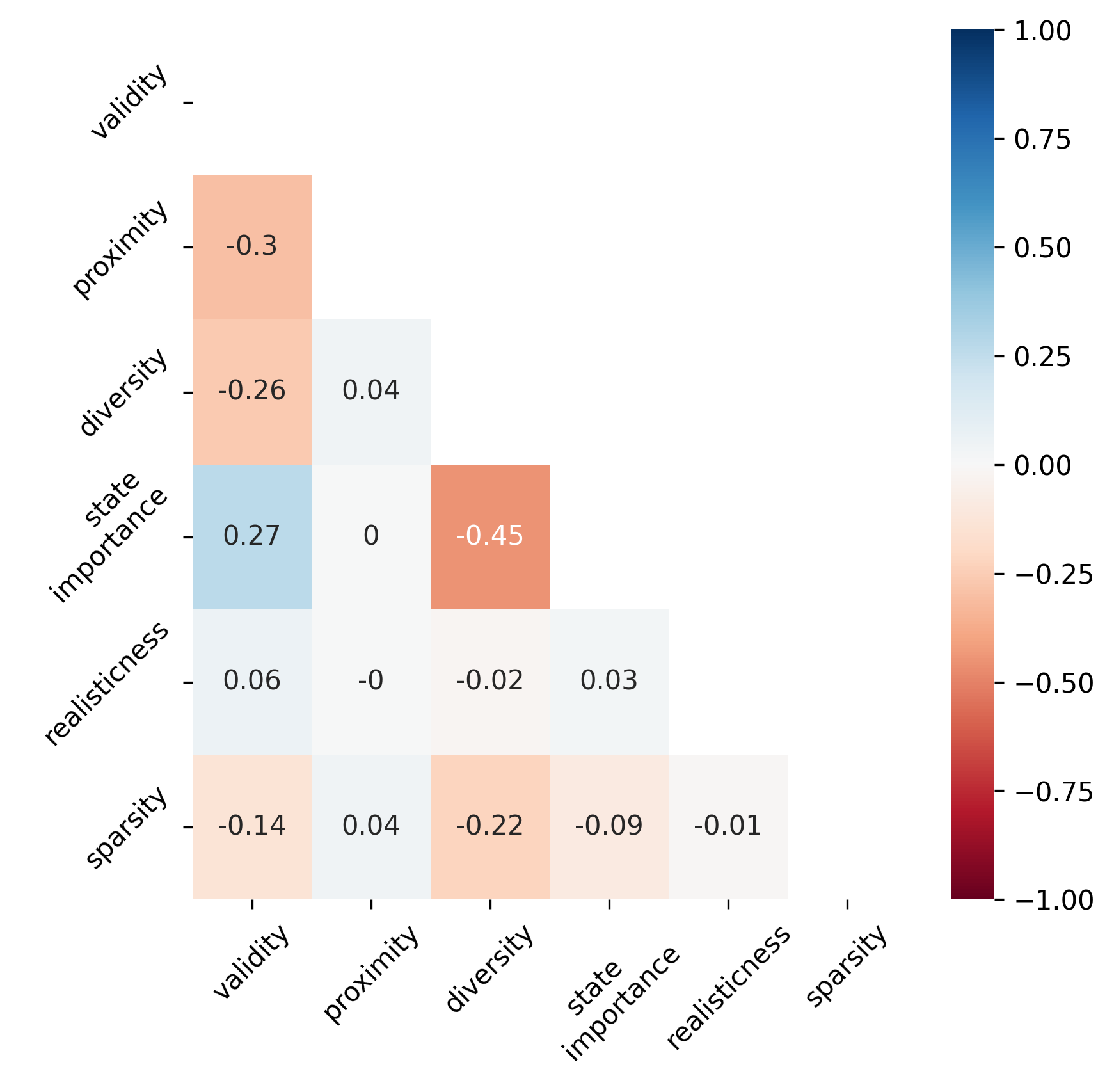}
    \caption{The average Pearson correlations between quality criteria of candidate CTEs in \textsf{\small DaC}.}
    \label{fig:qc_correlation}
\end{figure}

\textbf{Discussion: }
We hypothesise that the trade-off of \textsf{\small \textsf{\small Validity}} with \textsf{\small Sparsity} and \textsf{\small Proximity} appears because \textsf{\small \textsf{\small Validity}} pushes for longer and more dissimilar trajectories that can result in larger differences in rewards, while \textsf{\small Sparsity} pushes for shorter and \textsf{\small Proximity} for more similar trajectories.
However, \textsf{\small \textsf{\small Validity}} does synergise with \textsf{\small State Importance}, likely because both criteria prefer phases in the environment, where differences in rewards can be higher.
\textsf{\small Diversity} trades off against \textsf{\small State Importance} and \textsf{\small Sparsity}, because these criteria prefer certain start points and lengths of trajectories, while \textsf{\small Diversity} pushes for different start points and lengths. Possibly for the same reason, \textsf{\small Diversity} is negatively correlated with \textsf{\small \textsf{\small Validity}}.

Overall these results show that there are important trade-offs between quality criteria. To navigate these trade-offs it is important to carefully assign priorities between weights, that are adapted to the users' preferences. Further, these trade-offs pose a require the generation algorithms to find good compromises between the criteria.

\subsection{Performance of Generation Algorithms on Quality Criteria}
\label{app:qc-algorithms}
This Appendix provides the scores achieved on the individual quality criteria by each generation algorithm in Experiment 2 (see Section \ref{subsec:exp-2}).

\begin{figure*}[h]
  \begin{minipage}{0.8\textwidth}
    \centering
     \adjustbox{valign=t}{\includegraphics[width=\textwidth]{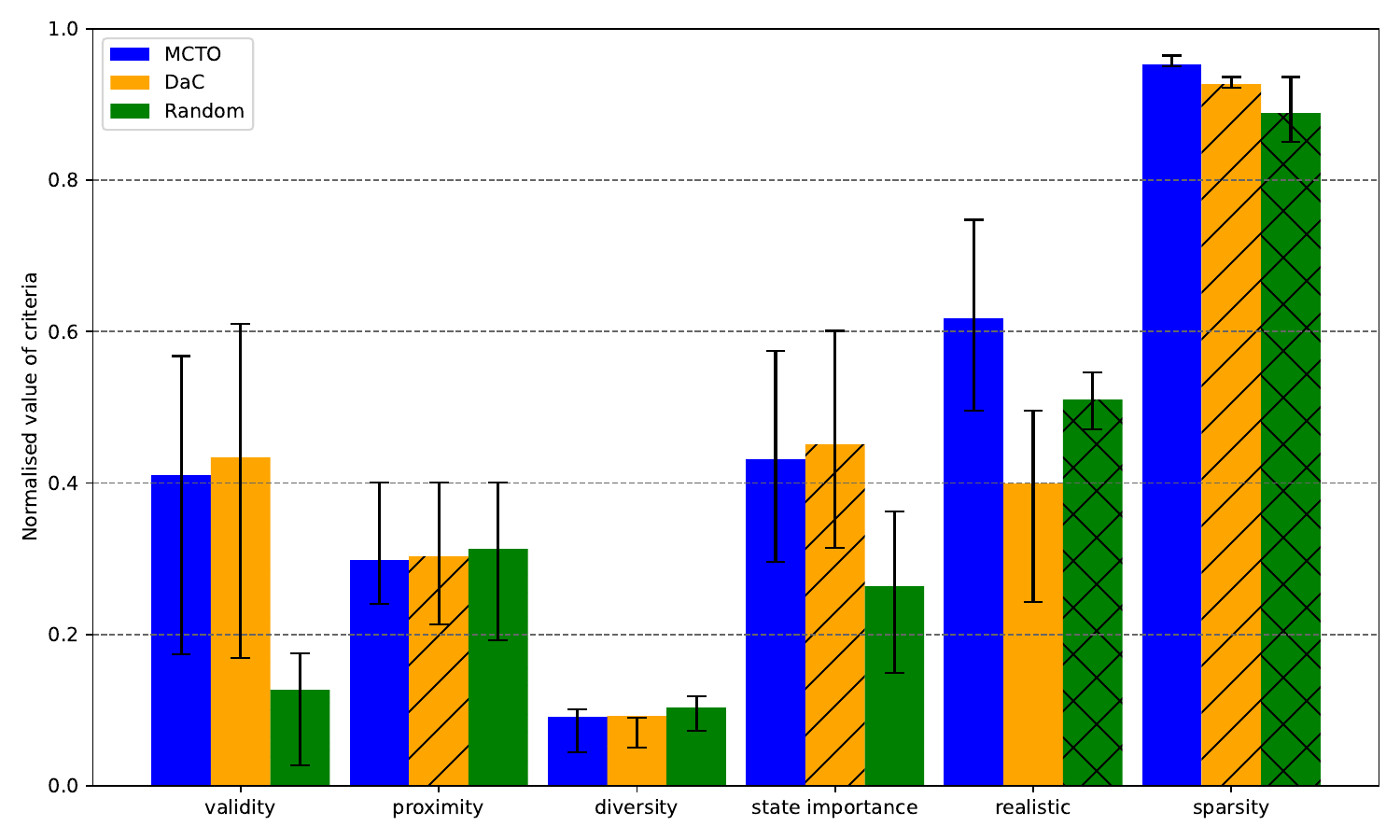}}
  \end{minipage}\hfill
  \begin{minipage}{0.2\textwidth}
    \centering
     \adjustbox{valign=t}{\includegraphics[width=\textwidth]{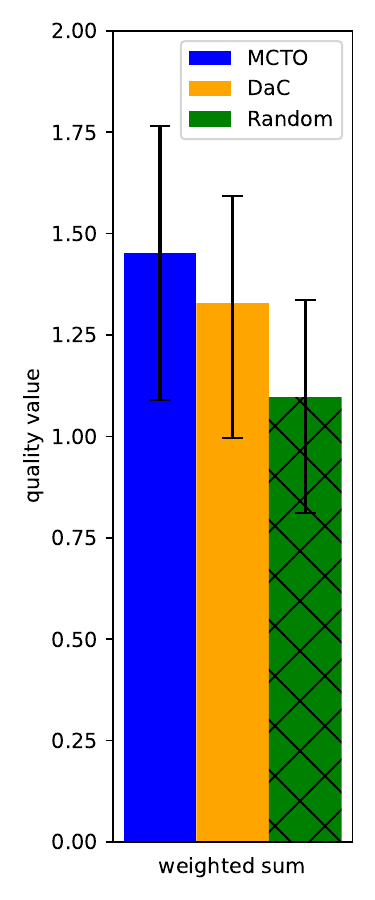}}
  \end{minipage}
\caption{For each quality criterion (left figure) and the quality value $\rho$ (right figure), the average normalised value and upper \& lower quartile achieved by the different generation methods are shown.}
\label{fig:qc-methods}
\end{figure*}

Figure \ref{fig:qc-methods} shows that \textsf{\small MCTO} achieved the highest average quality value, but did not perform best on every quality criterion.
The good performance of \textsf{\small MCTO} is largely attributed to its significantly higher scores on \textsf{\small Realisticness}, while \textsf{\small DaC} performed slightly better on \textsf{\small Validity} and \textsf{\small State importance}. Across most criteria, \textsf{\small MCTO} and \textsf{\small DaC} scored higher than \textsf{\small Random}.
This indicates that \textsf{\small MCTO} was best able to navigate the trade-offs between quality criteria allowing it to score high on the quality value.

It is also notable that the average values and ranges of quality criteria differed. \textsf{\small Sparsity} has very high averages but falls in a small range of values. Comparatively, \textsf{\small Diversity} has very low average values and the scores of \textsf{\small \textsf{\small Validity}} have a large range. 
This is a byproduct of the normalisation procedure. Each criterion is normalised to a range of $[0,1]$ through the highest and lowest scores recorded for that criterion.
While most trajectories are very short, thus scoring high on \textsf{Sparsity}, the longest ones can be very long. This causes most values to be very high for \textsf{Sparsity}. For the opposite reason, values of \textsf{Diversity} are normalised to be very low. For most generated CTEs there are a lot of previous CTEs, making it hard to be significantly different. However, there are some early CTEs which did score very high on \textsf{Diversity} and thus stretched the normalised range upwards.
Values for \textsf{Validity} do not fall in such a big range. Thus they are left with a bigger deviation after being normalised.

Importantly, a quality criterion with higher average values does not have a larger influence on which CTE. This is investigated in Appendix \ref{app:influence-exp}.

\subsection{Influence of quality criteria on the choice of CTE}
\label{app:influence-exp}
The previous section \ref{app:trade-offs-exp} indicated that there are trade-offs between quality criteria which need to be navigated in the multi-objective optimization problem of finding the CTE with the highest quality value. We want to find out whether some quality criteria have more or less influence on which CTE gets chosen.

\textbf{Experimental Setup:} Similarly to the Experiment in Appendix \ref{app:trade-offs-exp}, \textsf{\small DaC} is employed to generate 1 CTE for each of the 1000 random sets of weights $\upomega_{random}$. We record the quality criteria for the chosen CTE (the one that is shown to the user), but also for all candidate CTEs that were not chosen. Based on this we record for each criterion and each chosen CTE 1) how high the chosen CTE ranks amongst candidates according to that criterion (\textit{percentile ranking}), 2) the value of the criterion compared to the criterion's highest candidate CTE (\textit{relative value}) and 3) the correlation between the criterion and the quality values $\rho$ ($\rho$\textit{-correlation}).

\textbf{Results: }
\begin{table}[]
    \centering
\begin{tabular}{|l|ccc|}
\hline
 quality criteria   & percentile ranking $\uparrow$   & relative value $\uparrow$  &   $\rho$-correlation $\uparrow$ \\
                    & mean $|$ median        & mean $|$ median    &                  \\
                    \hline
 \textsf{\small \textsf{\small Validity}}           & \rule{0pt}{11pt} 79.6\% $|$ \textbf{95.9\%}      & 0.61 $|$ 0.68      &             \textbf{0.34} \\
 \textsf{\small Proximity}          & 55.7\% $|$ 64.4\%      & 0.9 $|$ 0.91       &             0.25 \\
 \textsf{\small Diversity}          & 36.6\% $|$ 28.8\%      & 0.37 $|$ 0.27      &            -0.1  \\
 \textsf{\small State Importance}& \textbf{82.7\%} $|$ 91.8\%      & 0.6 $|$ 0.58       &             0.32 \\
 \textsf{\small Realisticness}      & 69.7\% $|$ 93.2\%      & 0.9 $|$ 0.93       &             0.12 \\
 \textsf{\small Sparsity}           & 52.9\% $|$ 56.2\%      & \textbf{0.92} $|$ \textbf{0.96}      &             0.18 \\
\hline
\end{tabular}
    \caption{Statistics that describe how influential quality criteria are for determining the CTE in DaC. 1000 chosen CTEs are compared to their respective candidate CTEs and for each criterion we display on average how high the chosen CTE ranks, the value of the chosen CTE compared to the criterion's favorite and its correlation with the quality value.}
    \label{tab:combination-qc}
\end{table}
Looking at Table \ref{tab:combination-qc} it stands out that \textsf{\small \textsf{\small Validity}} and \textsf{\small State Importance} often get an option they rank very highly (better than $79.6\%$ and $82.7\%$ respectively), but that has a  much lower value relative to their favourite options ($0.61$ and $0.6$ times as high as their favourite). Their values do have the strongest correlation with the quality values ($0.34$ and $0.32$).
On the other hand, \textsf{\small Proximity}, \textsf{\small Realisticness} and \textsf{\small Sparsity} often have to content themselves with a CTE that they don't rank highly ($55.7\%$, $69.7\%$ and $52.9\%$), but get a value close to their highest rated CTE ($0.9$, $0.9$ and $0.92$ times as good as their favourite). Further, they have lower correlations ($0.25$, $0.12$ and $0.18$) with the quality values.
Lastly, \textsf{\small Diversity} scores lowest on all three measurements.

\textbf{Discussion: }
Table \ref{tab:combination-qc} shows us that criteria often did not get their most preferred outcome. This is shown by the fact that criteria mostly didn't get their top-ranked CTE chosen. This supports our conclusion from Appendix \ref{app:trade-offs-exp} that there are real trade-offs between the criteria, which means there is no easy choice which satisfies all criteria.

It is interesting that \textsf{\small \textsf{\small Validity}} and \textsf{\small State Importance} score high on percentile ranking and qc-correlation, but low on relative values, while \textsf{\small Proximity}, \textsf{\small Realisticness} and \textsf{\small Sparsity} show the opposite trend.
This phenomenon emerges because values decrease steeply when moving down the ranking of CTEs according to \textsf{\small \textsf{\small Validity}} and \textsf{\small State Importance}. Thus, CTEs that rank relatively high are already much worse than their highest-rated option. This means these criteria often don't achieve high absolute values. However, the large differences in values of these criteria cause large differences in the weighted sum. 
Thus they have an outsized impact on the quality value and the final decision. 
On the other hand \textsf{\small Proximity}, \textsf{\small Realisticness} and \textsf{\small Sparsity} have a less steep distribution of values, thus ``weaker opinions" about the choice of CTEs and end up influencing the weighted sum less.

It stands out that \textsf{\small Diversity} does not have a strong influence on the choice of CTE. This might be because \textsf{\small Diversity} is negatively correlated with multiple other criteria and thus gets crowded out when calculating the weighted sum. Furthermore, Appendix \ref{app:qc-algorithms} showed that \textsf{\small Diversity} has a small range of values, which might further contribute to its weak influence.

\subsection{Most and Least informative set of weights}
\label{subsec:best-worst}
\textbf{Experimental Setup:} To provide further insight into which quality criteria are important to create informative CTEs, we find the set of weights that lead to the most and least informative CTEs. 
Additionally, the most informative weights found here for each generation algorithm are also those used to generate CTEs in Experiment 1 (see Section \ref{subsec:exp-1}).

In Experiment 3 (see Section \ref{subsec:exp-3}) we test the informativeness of 30 sets of CTEs generated by 30 different sets of weights for the quality criteria $\omega \in \upomega = \{\omega_{Validity_j}, ..., \omega_{Sparsity_j}\}^{30}_{j=1}$. We use both \textsf{\small MCTO} and \textsf{\small DaC} to generate CTEs and train 10 $M_\phi$s on each set of CTEs. From that, we identify the best and worst performing set of weights $\omega$ on the single and contrastive task.
 
\begin{table*}[]
    \centering
    \begin{tabular}{|ll|cccccc|c|}
    \hline
         & $\omega$ & Validity & Proximity& Diversity & State Importance & Realisticness & Sparsity & \\
         \hline
 \multirow{2}{*}{contrastive} & best $\uparrow$ &\textbf{0.982} & \textbf{0.98}  & 0.576 &  0.528 &  0.303 &\textbf{0.851} & 
 \multirow{4}{*}{\rotatebox{90}{\textsf{\small MCTO}}}\\
 & worst $\downarrow$ &0.126 & 0.213 & 0.492 &  \textbf{0.943} &  0.052 &\textbf{0.752} & \\
 \cline{1-8}
 \multirow{2}{*}{single} & best $\uparrow$ &\textbf{0.878} & \textbf{0.92}  & \textbf{0.915} &  0.674 &  0.639 &0.657 & \\
 & worst $\downarrow$ &0.17  & 0.496 & 0.205 &  \textbf{0.968} & 0.203&\textbf{0.633} & \\
\hline
 \multirow{2}{*}{contrastive} & best $\uparrow$ &\textbf{0.902} & 0.658 & 0.405 &  0.587 &  0.12  &\textbf{0.988}  & \multirow{4}{*}{\rotatebox{90}{\textsf{\small DaC}}} \\
 & worst $\downarrow$ &0.006 & 0.267 & \textbf{0.891} &  \textbf{0.924} &  0.353 &0.579  & \\
 \cline{1-8}
 \multirow{2}{*}{single} & best $\uparrow$ &\textbf{0.878} & \textbf{0.92}  & \textbf{0.915} &  0.674 &  0.639 &0.657  & \\
 & worst $\downarrow$ &0.412 & \textbf{0.836} & 0.38  &  0.087 &  0.042 &\textbf{0.748}  & \\
\hline
    \end{tabular}
    \caption{The sets of weights $\omega$ (out of 30) that lead to the most and least informative CTEs for a Neural Network when optimised by \textsf{\small MCTO} or \textsf{\small DaC} for the contrastive and single task.}
    \label{tab:worst-best-mcts}
\end{table*}

\textbf{Results: }Table \ref{tab:worst-best-mcts} shows that the most informative CTEs generated by \textsf{\small MCTO} for an NN have high \textsf{Validity} and \textsf{Proximity}, while the least informative ones have high \textsf{State Importance} and \textsf{Sparsity}. For the most informative weights for contrastive $\omega_{Sparsity}$ was high, while $\omega_{Diversity}$ was high for single.

When generating highly informative CTEs with \textsf{\small DaC} the contrastive task $\omega_{Validity}$ and $\omega_{Sparsity}$ were high, while the least informative CTEs were generated with high $\omega_{Diversitiy}$ and $\omega_{State Importance}$. To score high on the single task, $\omega_{Validity}$, $\omega_{Proximity}$ and $\omega_{Diversity}$ were high, while $\omega_{Sparsity}$ and $\omega_{Proximity}$ were high for the lowest scores.

While \textsf{\small DaC} and \textsf{\small MCTO} agree on which set of weights leads to the most informative CTEs for the single task, they disagree on the three other measures.

\textbf{Discussion: }
These results again highlight the importance of \textsf{Validity} for creating CTEs that are informative to a Neural Network. Furthermore, they indicate that \textsf{State Importance} can be harmful. For \textsf{\small MCTO}, \textsf{Proximity} and \textsf{Diversity} are clearly helpful. On \textsf{\small DaC} there are less clear claims that can be made as \textsf{Proximity}, \textsf{Diversity} and \textsf{Sparsity} appear with high weights when informativeness was high and when it was low.
The fact that \textsf{\small MCTO} and \textsf{\small DaC} agree on only one out of four settings, indicates that different generation algorithms do not converge to the same priorities between criteria. However, since we only sample 30 weights these results might not be highly significant.

\section{Ablations and hyperparameters of generation algorithms}
\subsection{Ablations and hyperparameters of \textsf{\small MCTO}}
\label{app:ablation-mcto}
In \textsf{\small MCTO} there are multiple hyperparameters to be adjusted:

\begin{itemize}
    \item Number of starting points $n_{starts}$: \textsf{\small MCTO} is started at a specific point $s_n$ of the original trajectory. We rerun \textsf{\small MCTO} $n_{starts}$ times for different starting states $s_n \in \tau_{org}$ and select the CTE with the highest quality value. The $n_{starts}$ starting points are selected as the $n_{start}$ states which are rated highest by the \textsf{\small State Importance} criterion.
    \item Number of iterations $n_{iterations}$: From a given state \textsf{\small MCTO} starts to explore the tree through selection, expanding, simulation and back-propagating. $n_{iterations}$ denotes the number of iterations of these four steps in a state before a decision for the next action has to be made.
    \item Likelihood to terminate $p_{MCTO}(end)$: During each step of the simulation the trajectory is ended with a likelihood of $p_{MCTO}(end)$.
    \item Action threshold $threshold_a$: As a heuristic, we do not consider every action as a possible branch in the tree, but only those actions $a$ that are sufficiently likely according to the policy so that $\pi_\theta \geq threshold_a$.
    \item Discount factor $\gamma$: During backpropagation, a factor $\gamma$ can be employed to discount the rewards at each node. To switch discounting off the discount factor is set to $\gamma=1$.
\end{itemize}

Furthermore, we experiment with two heuristics:
\begin{itemize}
    \item Expansion heuristic: When deciding which branch to extend, we can choose through uniform sampling (\textit{random-expansion}) or according to a heuristic that takes into account an early estimate of the quality value (\textit{heuristic-expansion}). This heuristic tries each possible action, assumes the resulting trajectory as a leaf node and calculates the quality value of the resulting CTE. Then the action which achieved the highest quality value is chosen.
    \item Action selection during simulation: During the simulation, we can sample actions uniformly (\textit{random-simulation}) or according to the policy $\pi_\theta$ (\textit{policy-simulation}).
\end{itemize}

\subsubsection{Experimental Setup:}
We test all different options by using \textsf{\small MCTO} to generate 100 CTEs each for a set of weights $\upomega=\{\omega_{Validity_j}, ..., \omega_{Sparsity_j}\}^{100}_{j=1}$ that is uniformly sampled  $\omega_i \sim U(0,1)$. We measure the quality criteria and report on their quality values to compare options. 
We keep the following values for all experiments and only alter the parameter in question: $threshold_a=0.003$, $p_{MCTO}(end)=0.2$, $\gamma=0.85$, $n_{iterations}$, $=10$, \textit{heuristic-expansion}, \textit{policy-action} and $n_{starts}$ $=2$.
Aside from altering the hyperparameters and measuring the resulting quality values, we also record the efficiency as the average time needed to generate a CTE ($\frac{seconds}{CTE}$).

\subsubsection{Results:}
\begin{figure}
  \begin{minipage}{0.47\textwidth}
    \centering
    \adjustimage{height=5cm,valign=t}{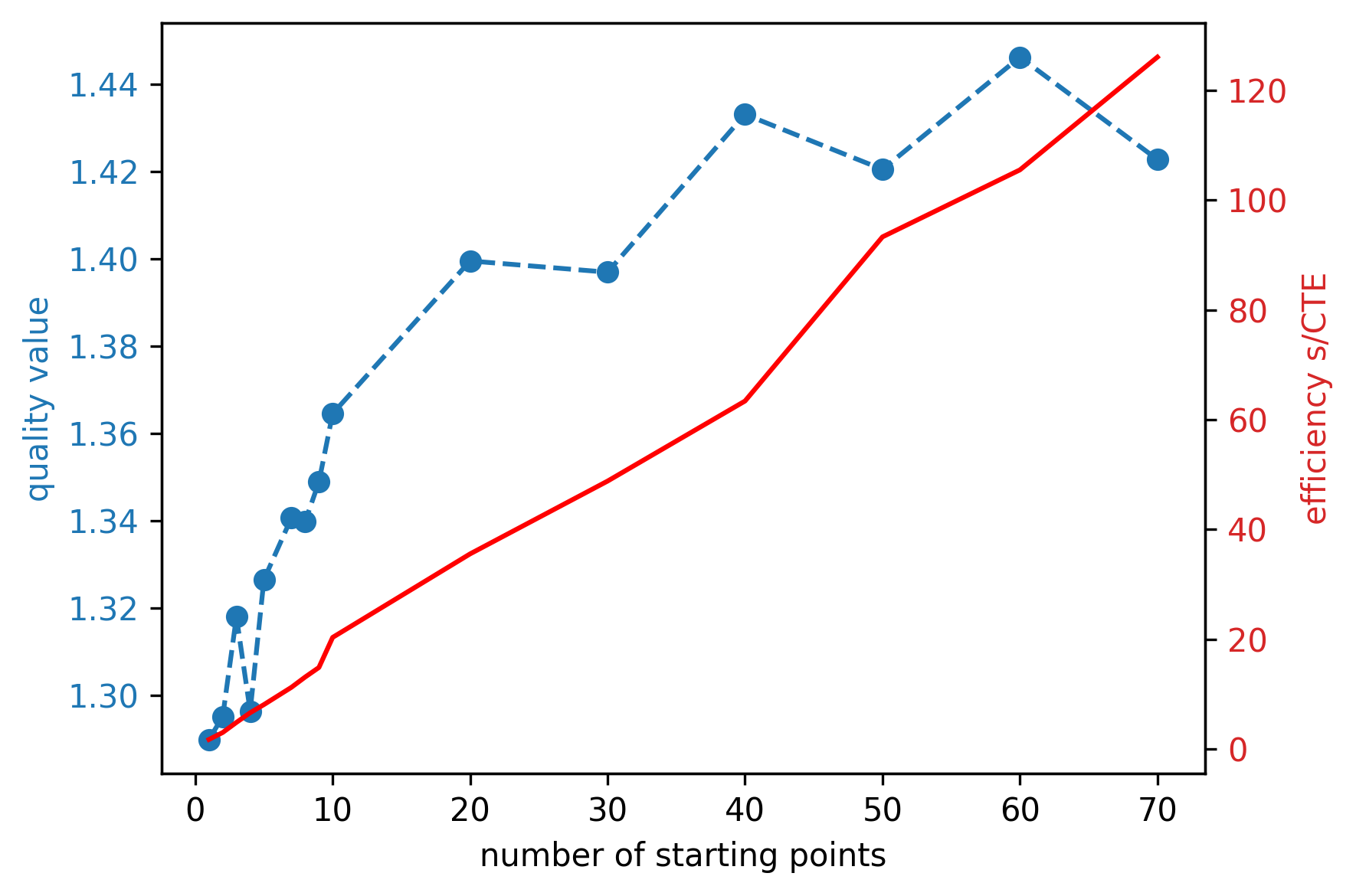}
    \caption{Average quality values of quality criteria for CTEs created by \textsf{\small MCTO} with differing number of starting points along with the efficiency in $\frac{seconds}{CTE}$.}
    \label{fig:num-starts}
  \end{minipage}\hfill
  \begin{minipage}{0.47\textwidth}
    \centering
    \adjustimage{height=5cm,valign=t}{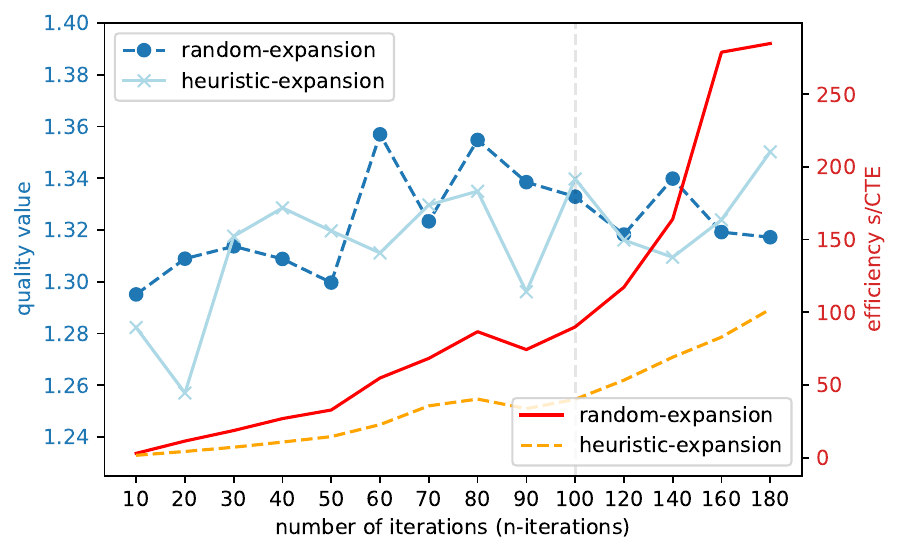}
    \caption{Avg. Quality Value (left y-axis and blue graphs) for CTEs generated by \textsf{\small MCTO} and efficiency of generation (right y-axis with red graphs) using different numbers of iterations across a setting with (\textit{heuristic-expansion}) and without (\textit{random-expansion}) the expansion heuristic.}
    \label{fig:num_iters}
  \end{minipage}
\end{figure}

\textbf{Number of starting points:}
Figure \ref{fig:num-starts} shows how the quality value of CTEs increases when we choose from more candidate CTEs produced by  \textsf{\small MCTO} from a larger number of starting points.

It displays a clear trend that increasing the number of starting points increases the quality of the resulting CTE. Further, the graph seems to be convex with some noise in measurements. This makes sense since having more different options to select the best from is strictly better, but has diminishing returns when considering more options.
However, this comes at the cost of efficiency, with the time needed to compute a CTE growing linearly with the number of starting points. 

Despite the increased computing time and diminishing returns we deem the increase in performance as significant. Thus we choose to run \textsf{\small MCTO} for every possible starting state, setting $n_{starts}=70$ for our experiments.

\textbf{Number of iterations:}
We test multiple amounts of iterations per step for a setting using ``random-expansion" and ``qc-expansion". Figure \ref{fig:num_iters} shows a slight improvement in quality values with increasing $n_{iterations}$. 
This is backed by the small correlation between $n_{iterations}$and quality values of $0.092$ ($p=0.01$) for \textit{random-expansion} and $0.04$ ($p=0.18$) for \textit{heuristic-expansion}.
However, with every additional iteration a new simulation needs to be run leading to a roughly linear decrease in efficiency for \textit{random-expansions}. 

This presents us with another trade-off between quality and computational efficiency. Since the benefits of increasing the number of iterations are less clear, we choose $n_{iterations}=10$ for our experiments.

\textbf{Likelihood to terminate:}
Values between $[0.05, ..., 1.0]$ are used as the likelihood of the simulation to terminate in a given step.

\begin{figure}
    \centering
    \includegraphics[height=5cm]{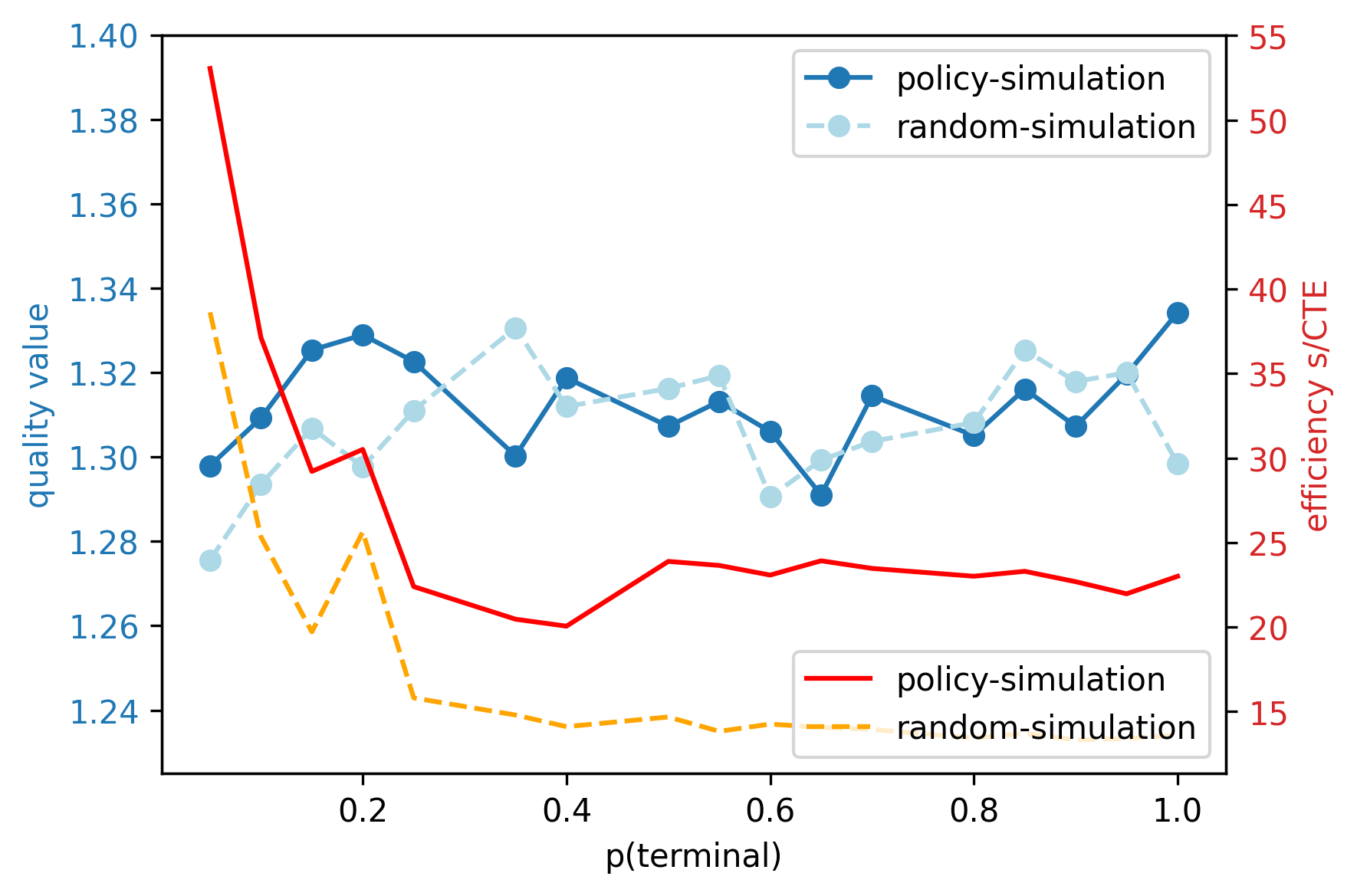}
    \caption{Avg. quality value for CTEs generated by \textsf{\small MCTO} for different likelihoods of termination during the simulation $p_{MCTO}(end)$. We show this for \textit{policy-} and \textit{random-simulation} along with the efficiency in $\frac{seconds}{CTE}$.}
    \label{fig:p_term}
\end{figure}

Figure \ref{fig:p_term} shows no clear trend in quality values when increasing $p_{MCTO}(end)$. One can make out an increase in the early values ($0.05 - 0.2$) and a dip around $p_{MCTO}(end)=0.65$. However, these observations might be measurement inaccuracies rather than consistent trends. 
For low likelihoods of ending ($p_{MCTO}(end) \leq 0.2$) there are significant increases in the computational cost for generating CTEs. This is because individual simulations become longer. 

For our experiments, we decide to use $p_{MCTO}(end)=0.35$ since it achieves the highest quality value in combination with \textit{random-simulation} and is reasonably efficient.

\begin{figure}
  \begin{minipage}{0.45\textwidth}
    \centering
     \adjustbox{valign=t}{\includegraphics[height=5cm]{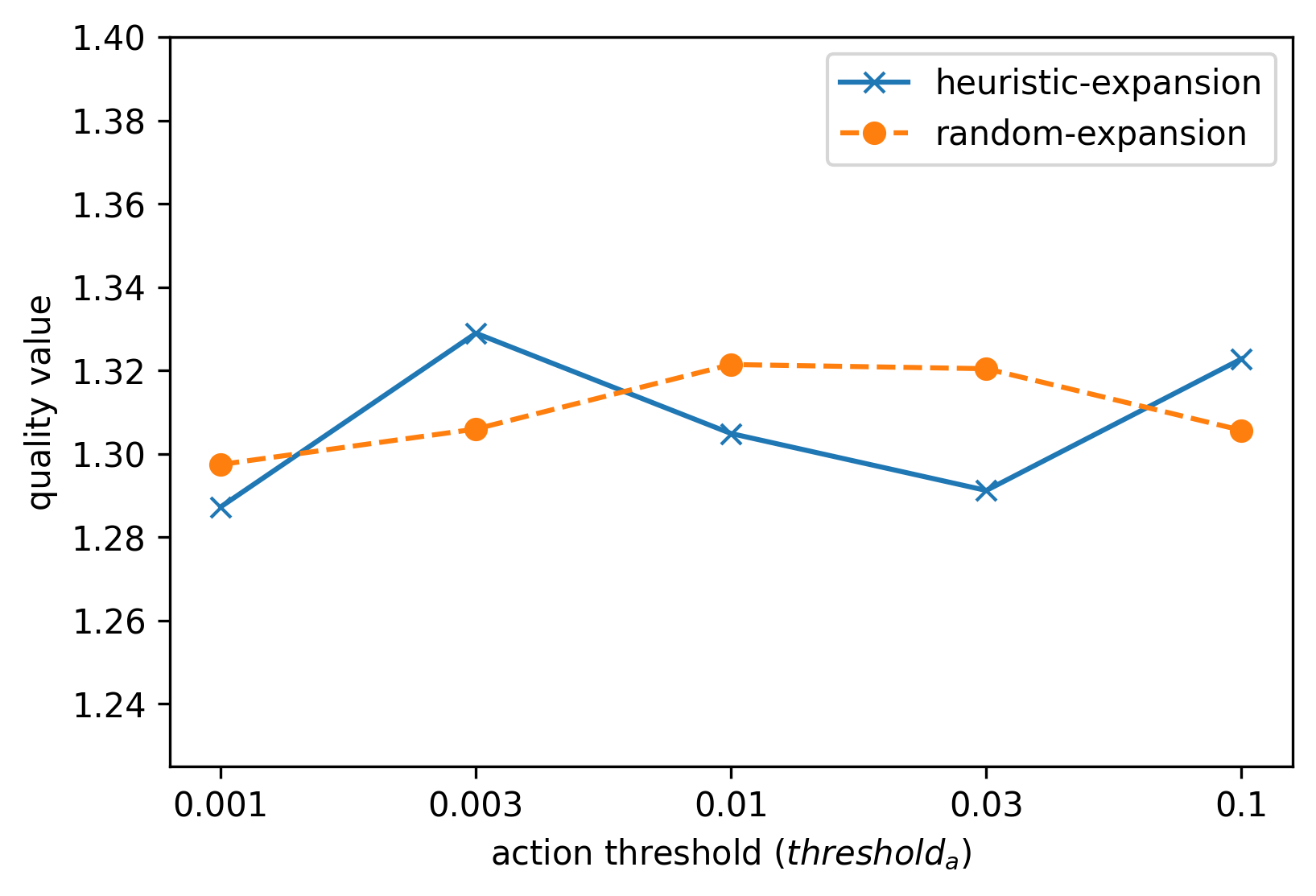}}
    \caption{Avg. quality value for CTEs generated by \textsf{\small MCTO} for different action thresholds $threshold_a$ using both \textit{random-} and \textit{heuristic-expansion}.}
    \label{fig:action_threshold}
  \end{minipage}\hfill
  \begin{minipage}{0.45\textwidth}
    \centering
     \adjustbox{valign=t}{\includegraphics[height=5cm]{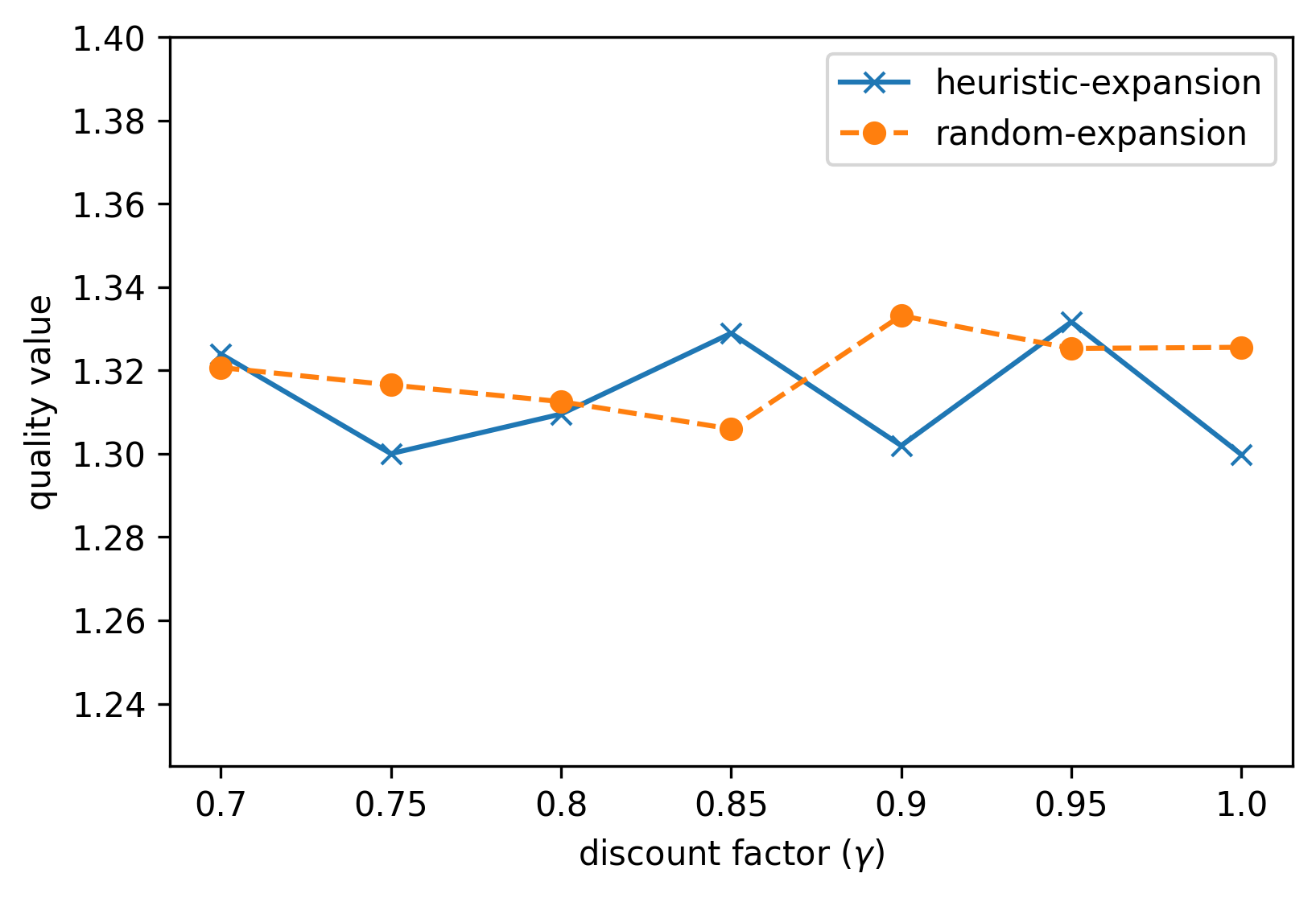}}
    \caption{Avg. quality value for CTEs generated by \textsf{\small MCTO} for different discount factors $\gamma$ using both \textit{random-} and \textit{heuristic-expansion}.}
    \label{fig:discount_factor}
  \end{minipage}
\end{figure}

\textbf{Action threshold:}
We try a range of values for $threshold_a \in \{0.001, 0.003, 0.01, 0.03, 0.1\}$. There is no clearly best $threshold_a$ recognisable from Figure \ref{fig:action_threshold}. None of the differences between the action thresholds is significant and it can be concluded that this hyperparameter is not important for the functioning of \textsf{\small MCTO}. For our experiments, we choose $threshold_a=0.03$.

\textbf{Discount factor:}
We test values for the discount factor between $[0.7, .., 1.0]$ applied to the result of a simulation.

Figure \ref{fig:discount_factor} does not show a significant impact of the discount factor on the quality value of resulting CTEs. There is also no significant correlation linking the discount factor and quality value. We conclude that adding a  discount factor does not help or hinder \textsf{\small MCTO} in its search and thus choose $\gamma=1.0$ for simplicity.

\textbf{Expansion Heuristic:}
To understand how the method for choosing which branch to extend influences the quality value of the resulting CTEs, we deploy \textit{heuristic-expansion} and \textit{random-expansion} across different values for $threshold_a$, $\gamma$ and $n_{iterations}$ and average the results.

\begin{table}[]
    \centering
    \begin{tabular}{|c|rrr|}
    \hline
         &  $threshold_a$ &$\gamma$ &$n_{iterations}$\\
         \hline
         heuristic-expansion& 
    1.307 &1.314 &1.315\\
 random-expansion&1.310 &1.320 &1.323\\
 \hline
 p-value&0.912 &0.793 &0.640\\
 \hline
 \end{tabular}
    \caption{The average quality value of CTEs generated by \textsf{\small MCTO} using \textit{heuristic-expansion} or \textit{random-expansions} across multiple values from $threshold_a$, $\gamma$ and $n_{iterations}$ respectively, along with the p-value indicating the statistical significance in differences between them.}
    \label{tab:expansion}
\end{table}

For all three test cases, \textit{random-expansion} slightly outperforms \textit{heuristic-expansion}. However, none of the differences are statistically significant.

This indicates that an early estimation of the quality criteria is not a good heuristic to guide the expansions in the search tree. This might be because the early estimations do not accurately reflect how promising a branch is to extend. Furthermore, using the heuristic comes at a significant computational cost. Across all $n_{iterations}$ \textit{heuristic-expansion} needs $61.3 \frac{s}{CTE}$, while \textit{random-expansion} only takes $23.2 \frac{s}{CTE}$ (see Figure \ref{fig:num_iters}). Thus we choose to use \textit{random-expansion} for our experiments.

\textbf{Action selection during simulation:}
We compare the two proposed methods for selecting actions during the simulation, \textit{policy-simulation} and \textit{random-simulation}, across different values for $p_{MCTO}(end)$ and average the quality values of the resulting CTEs.

There is no statistically significant difference between the two methods with \textit{policy-simulation} averaging a quality value of $1.295$ and \textit{random-simulation} achieving $1.302$. 
Additionally, Figure \ref{fig:p_term} shows us that following the policy is less efficient than choosing randomly. Thus we decide to use \textit{random-simulation} for our experiments.

\subsection{Ablations of Deviate and Continue}
\label{app:ablation-dac}
We ablate three aspects of \textsf{\small DaC}:

\begin{enumerate}
    \item Number of deviations: The counterfactual trajectory can either perform one (\textit{"1-deviation"}) or multiple (\textit{"n-deviations"}) deviations. We use the same process for later deviations as for the first deviation. An action is chosen according to the policy, but actions which would lead the counterfactual to meet the original trajectory are excluded.
    \item Action selection: After deviating from the original trajectory the actions in the counterfactual trajectory can either be chosen by following the policy $\pi_\theta$ (\textit{"continue-policy"}) or by uniformly sampling random actions (\textit{"continue-random"}).
    \item Likelihood to terminate: At each step of the continuation phase the trajectories have a chance $p_{DaC}(end)$ of ending.
\end{enumerate}

\subsubsection{Experimental Setup:}
For each setting \textsf{\small DaC} is used to generate 200 CTEs each for a set of weights $\upomega=\{\omega_{Validity_j}, ..., \omega_{Sparsity_j}\}^{200}_{j=1}$ that is uniformly sampled  $\omega_i \sim U(0,1)$. 
We test the combination of values for \textit{n-deviations} $\in \{1,2,3\}$ and $p_{DaC}(end)\in\{0.05,0.1, ..., 1.0\}$. Furthermore, we compare \textit{continue-policy} and \textit{continue-random}, while using ``1-deviation'' and $p_{DaC}(end)=0.5$.

\subsubsection{Results:}
\begin{table}[ht]
\centering
\begin{tabular}{|l|rrr|}
\hline
 & 1-deviation & 2-deviations & 3-deviations\\
\hline 
avg. quality value & 1.339 & 1.334 & 1.346 \\
\hline
\end{tabular}
\caption{Average quality value achieved across all values for $p_{DaC}(end)$ for different numbers of deviations.}
\label{tab:deviations}
\end{table}

\begin{figure}
  \begin{minipage}{0.67\textwidth}
    \centering
     \adjustbox{valign=t}{\includegraphics[height=5cm]{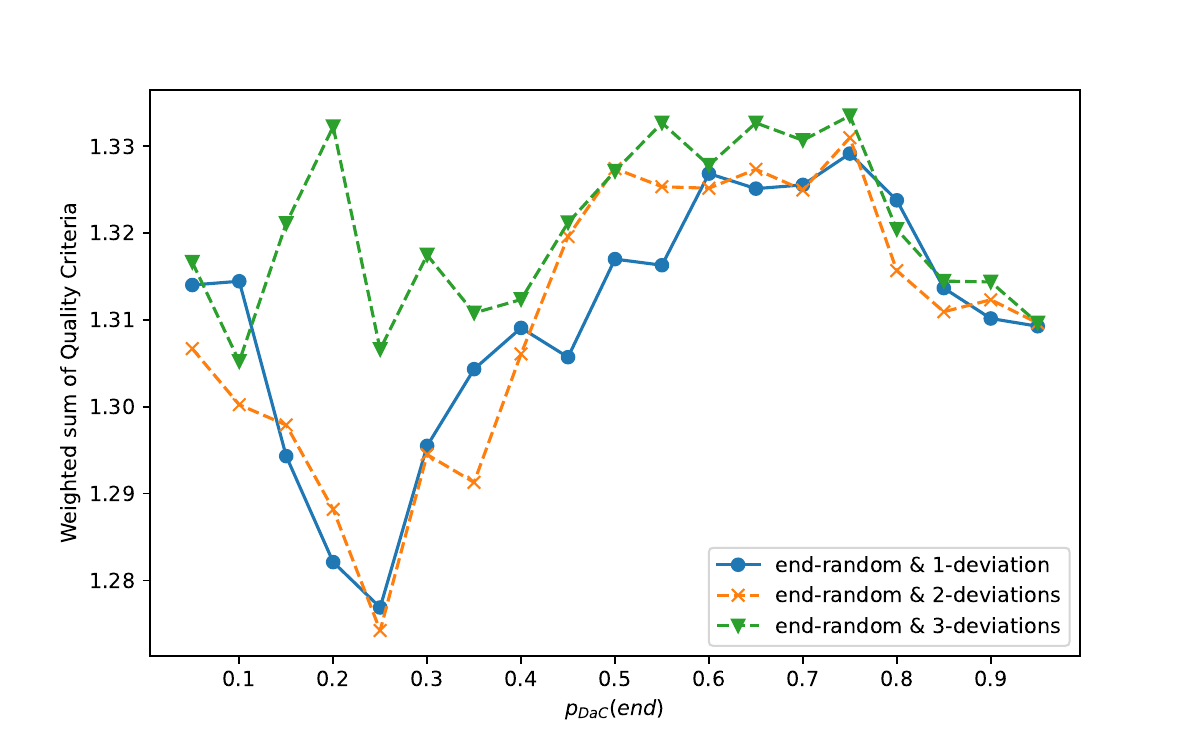}}
    \caption{Quality values achieved by different methods for different values of $p_{DaC}(end) \in \{0.05, ... 0.95\}$.}
    \label{fig:ending_methods_step}
  \end{minipage}\hfill
  \begin{minipage}{0.3\textwidth}
    \centering
     \adjustbox{valign=t}{\includegraphics[height=5cm]{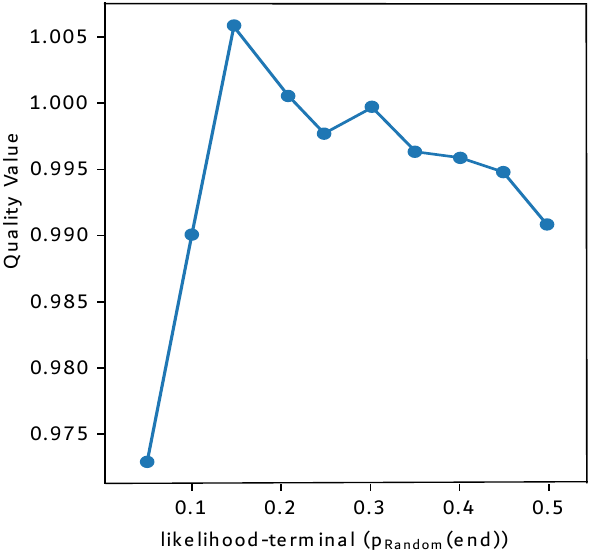}}
    \caption{Average Quality Value achieved by \textsf{\small Random} with different likelihoods of termination}
    \label{fig:random_p_term}
  \end{minipage}
\end{figure}

\textbf{Number of deviations: }Table \ref{tab:deviations} shows that the differences in quality values between different amounts of deviation steps are small. In fact, the differences are statistically not significant indicating that the number of deviations is not essential.  However, from Figure \ref{fig:ending_methods_step} we can make out that ``3-deviations" outperforms methods with fewer steps on most values for $p_{DaC}(end)$. This is largely because ``3-deviations" regularly achieve higher values for \textsf{Proximity} ($p<0.05$). The highest values are achieved by "3-deviations" between $0.5 \leq p_{DaC}(end) \leq 0.75$. We thus choose to use ``3-deviations''.

\textbf{Action selection: }Averaged over multiple values of $p_{DaC}(end)$, continuing with the policy achieves an average quality value of $1.347$, while \textit{random-continuation} achieved $1.351$. However, these results are not statistically significant. Since \textit{policy-continuation} led to more informative CTEs according to the researchers' subjective judgement, we choose to use \textit{policy-continuation} in our experiments.

\textbf{Likelihood to terminate: }Figure \ref{fig:ending_methods_step} shows a trend that high-quality values are achieved for low $p_{DaC}(end)$ values. As $p_{DaC}(end)$ rises, quality values first drop until $p_{DaC}(end)=0.25$ and then steadily increase to a maximum of around 0.7, before dropping off again. Moderately high values score well because they produce short CTEs that are rated high on \textsf{Sparsity}. Lower values perform well on \textsf{Diversity} because they can lead to a wide range of lengths of CTEs. 

For our experiments, we decide to choose "3-deviations" with $p_{DaC}(end)=0.55$.

\subsection{Hyperparameters of Random CTE Generation}
\label{app:ablation-random}
To make \textsf{\small Random} a stronger baseline comparison we perform a hyperparameter search to find the best way of choosing when to end the random trajectory.

\subsubsection{Experimental Setup:}
Similarly to \textsf{\small DaC} and \textsf{\small MCTO}, at every step in the trajectory \textsf{\small Random} has a likelihood of $p_{Random}(end)$ ending the counterfactual and original trajectory. For multiple values of $p_{Random}(end)$, 1000 CTEs are generated for 1000 randomly sampled sets of weights and their quality values $\rho$ are measured.

\subsubsection{Results}
Figure \ref{fig:random_p_term} shows a clear trend where the average quality values achieved by \textsf{\small Random} first rise until $p_{Random}(end)=0.15$ and then steadily fall. However, $p_{Random}(end)$ does not strongly influence the performance of \textsf{\small Random} with quality values only differing in a range of $<0.04$. Consequently, we choose $p_{Random}(end)=0.15$ for our experiments.

\section{Ablation for informativeness of quality criteria using DaC}

\label{app:dac-informativeness}
\textbf{Experimental Setup: }
The same experimental setup as described in Experiment 3 of the main paper (see Section \ref{subsec:exp-3}) is used to evaluate the influence of the weights $\omega$ of quality criteria for the informativeness of CTEs. In contrast to section \ref{subsec:exp-3} the CTEs are generated by \textsf{\small DaC} instead of \textsf{\small MCTO}.

\begin{figure}
    \centering
    \includegraphics[width=0.5\textwidth]{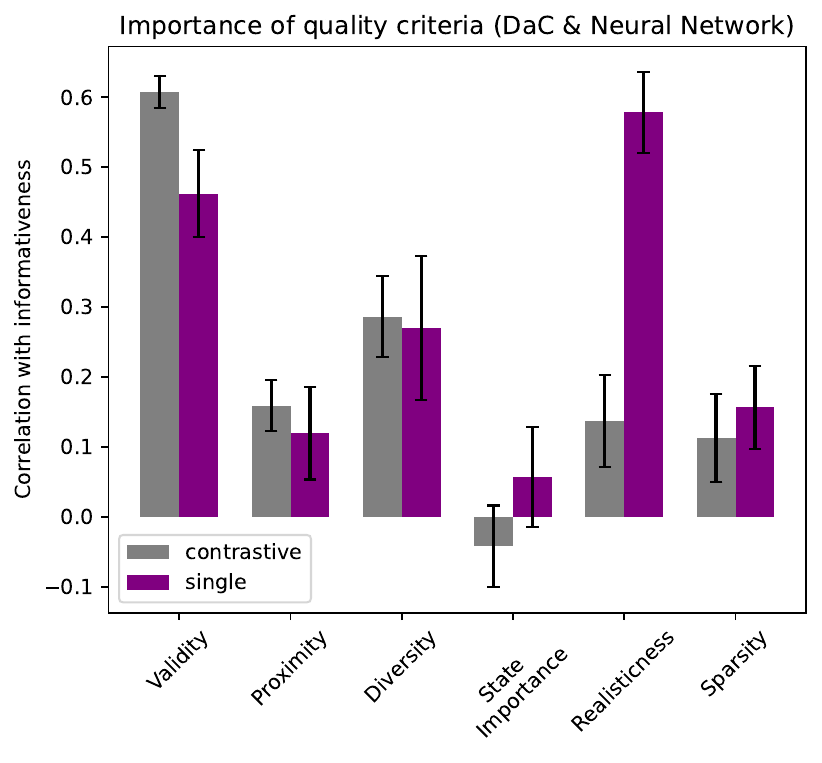}
    \caption{Spearman correlation between weights for the quality criteria and the informativeness of the CTEs generated by \textsf{\small DaC} for $M_\phi$ for the contrastive and single task. Averaged over 10 models along with the median and upper and lower quartile.}
    \label{fig:weight-correlation-dac}
\end{figure}

\textbf{Results: }
The weights for \textsf{\small Validity} $\omega_{Validity}$ correlate the strongest with contrastive performance and also strongly with single performance. $\omega_{Realisticness}$ stands out as highly correlated with single, despite being only slightly correlated with contrastive. Higher weights for \textsf{\small Diversity} also indicate higher performance on both tasks. $\omega_{State Importance}$ is not strongly correlated with both task performances.

\textbf{Discussions: }
The results for \textsf{\small DaC} present notable differences from those for \textsf{\small MCTO} (see Experiment 3 in section \ref{subsec:exp-3}). Relatively, \textsf{\small Realisticness} and \textsf{\small Sparsity} are more beneficial for making informative CTEs for \textsf{\small DaC} and \textsf{\small Proximity} and \textsf{\small State Importance} are less important than for \textsf{\small MCTO}.
However, there are also similarities. \textsf{\small Validity} is very important for both tasks, and the correlations for \textsf{\small Diversity} are similar for the two settings.

These results can increase our confidence that \textsf{\small Validity} and \textsf{\small Diversity} are generally important criteria for a Neural Network to learn from. The results also show that the priority between quality criteria does depend on the specific generation algorithm that is used. This might be because different generation algorithms have different ways of navigating the trade-offs between quality criteria.

\section{Ablation using Linear Model as proxy-human model}
\label{sec:linear-model}

In the main experiments, we use a Neural Network as a proxy-human model $M_\phi$ that receives and learns from the CTEs. To test whether our results are robust to using a different type of explainee we rerun the experiments using a Linear Regression Model (LM). If results are similar this can give us some indication that they are not only specific to a Neural Network but might be generally applicable to other explainees.

\subsection{Importance of Quality Criteria for a Linear Model}
\label{app:qc-lm}
This experiment aims to test whether the previous results about the importance of quality criteria hold up when using Linear Regression Models (LMs) instead of a Neural Networks (NNs) as a proxy-human model $M_\phi$. 

It repeats the Experimental Setup described in Experiment 3 (see Section \ref{subsec:exp-3}), except that $M_\phi$ is represented via LMs instead of NNs. Specifically, we use Simple Linear Regression Models with a bias term trained via regression.
Instead of performing multi-task learning, as we did with the Neural Network, we simply train two independent Linear Models on the single and contrastive task respectively.

\textbf{Experimental Setup: }
We repeat the experimental setup described in Experiment 3 (see Section \ref{subsec:exp-3}), except that $M_\phi$ is represented via LMs instead of NNs. Specifically, we use Simple Linear Models with a bias term trained via regression.
Instead of performing multi-task learning, as we did with the Neural Network, we simply train two independent Linear Models on the single and contrastive task respectively.
The Linear Models was trained with learning rate$=0.1$ and weight decay regularisation of $0.01$.

\begin{figure}
  \begin{minipage}{0.47\textwidth}
    \centering
     \adjustbox{valign=t}{\includegraphics[width=\textwidth]{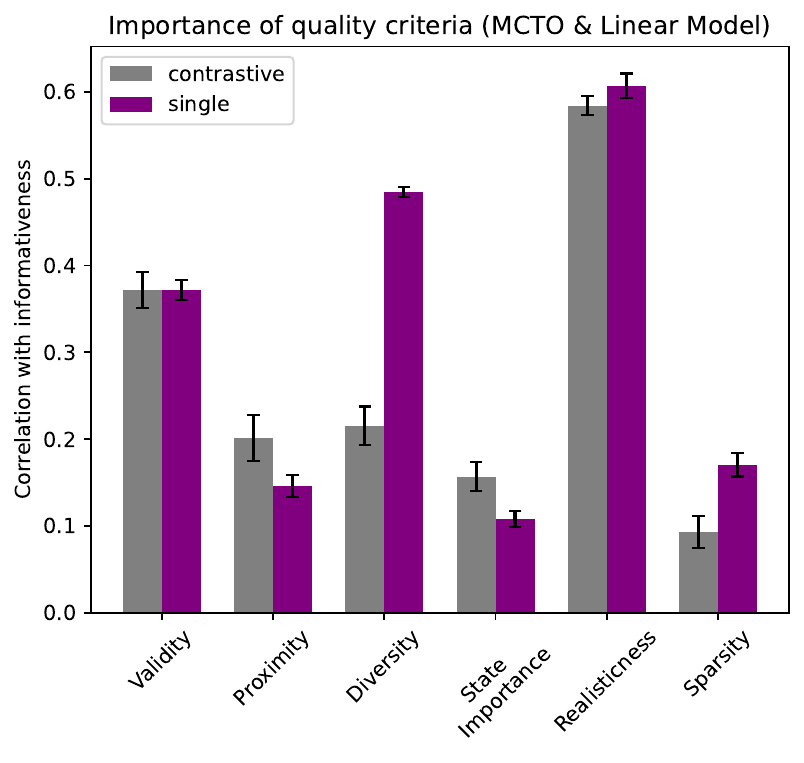}}
    \caption{Spearman correlation between the weights of quality criteria and the performance of the Linear Regression Model trained on CTEs generated by \textsf{\small MCTO} on the contrastive and single task averaged over 10 seeds.}
    \label{fig:weight-linear-mcts}
  \end{minipage}\hfill
  \begin{minipage}{0.47\textwidth}
    \centering
     \adjustbox{valign=t}{\includegraphics[width=\textwidth]{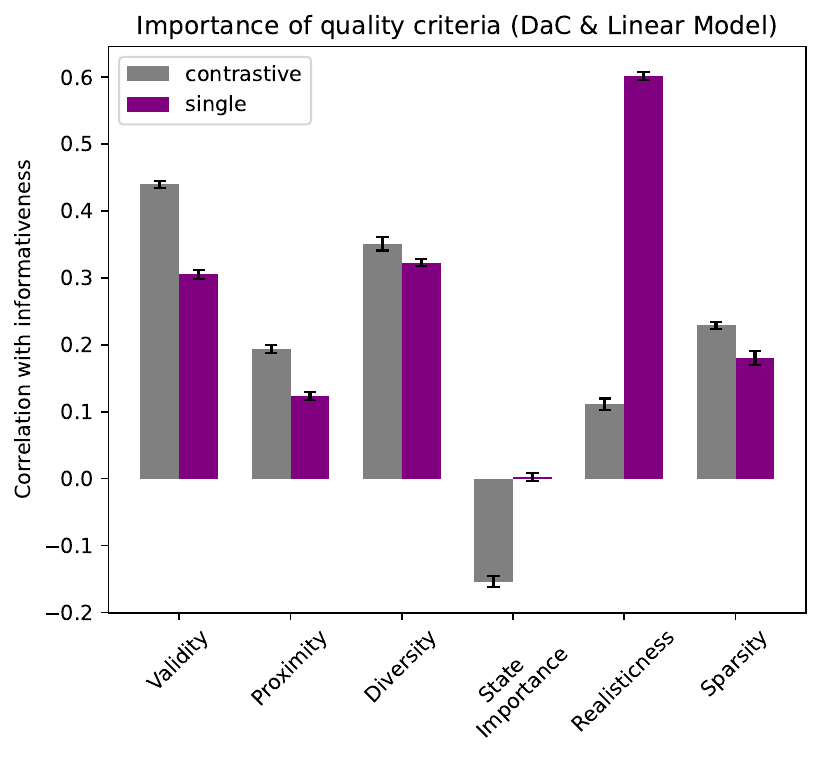}}
    \caption{Spearman correlation between the weights of quality criteria and the performance of the Linear Regression Model trained on CTEs generated by \textsf{\small DaC} on the contrastive and single task averaged over 10 seeds.}
    \label{fig:weight-linear-dac}
  \end{minipage}
\end{figure}

\textbf{Results: }
For the Linear Model trained on CTEs generated by \textsf{\small MCTO}, \textsf{\small Realisticness} was the most important criterion for both tasks (see Figure \ref{fig:weight-linear-mcts}). Furthermore, it stands out that \textsf{\small Diversity} is especially important for single, while \textsf{\small Validity} is important for both tasks. For all other criteria, their weights are slightly positively correlated with informativeness to the proxy-human Linear Model.

Similarly, for the informativeness of CTEs generated by \textsf{\small DaC} Figure \ref{fig:weight-linear-dac} shows that \textsf{\small Realisticness} is the most important criterion for single, while the weights for \textsf{\small Validity} $\omega_{Validity}$ correlate strongest with informativeness on the contrastive task. $\omega_{Diversity}$ proves important for both tasks, while $\omega_{State Importance}$ is negatively correlated with informativeness for contrastive.

\textbf{Discussion: }
Overall the results for the importance of quality criteria for informativeness are similar when using a Linear Model or a Neural Network. 

Notable differences for \textsf{\small MCTO} are that \textsf{\small Realisticness} is considerably more important for an LM on both tasks and \textsf{\small Diversity} is more important for single. Furthermore, \textsf{\small State Importance} and \textsf{\small Sparsity} now have a slight correlation with performance. \textsf{\small Validity} is relatively less important for the Linear Model than for the Neural Network.
For \textsf{\small DaC} the importance of quality criteria for informativeness is very similar for the Linear Model and Neural Network.

Overall there are some differences in importance when using \textsf{\small MCTO} and very few for \textsf{\small DaC}. This can give us more confidence that the importance of different quality criteria for generating informative CTEs is similar for different explainees.
However, we cannot simply extrapolate these results to all possible explainees but need to reevaluate which criteria should take priority. This applies especially when showing CTEs to humans instead of ML models.

\subsection{Informativeness of Explanations for a Linear Model}
It's unclear whether CTEs are especially informative when using a Neural Network as a proxy-human model or whether less complex architectures benefit equally from CTEs. Thus we ablate Experiment 1 (see Section \ref{subsec:exp-1}) by using Linear Models instead of Neural Networks as the proxy-human models $M_\phi$.

\textbf{Experimental Setup: }
We repeat the experimental setup from Experiment 1 (see section \ref{subsec:exp-1}). We train the Linear Models as described in Appendix \ref{app:qc-lm}.

\begin{figure}[h!]
    \centering
    \includegraphics[width=0.41\textwidth]{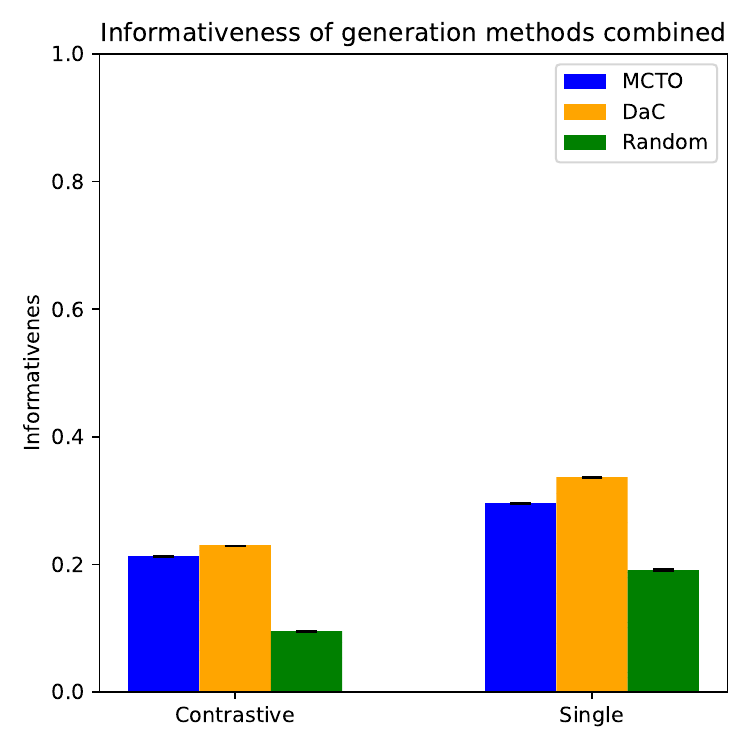}
    \label{fig:methods_informativeness_combinedlm}
    \caption{The informativeness of CTEs generated by \textsf{\small MCTO}, \textsf{\small DaC} and \textsf{\small Random} for a Linear Model on the single and contrastive task.}
\end{figure}

\textbf{Results: }
\textsf{\small DaC} performs best on both tasks, while \textsf{\small MCTO} outperforms \textsf{\small Random}. Overall informativeness of CTEs is low for the LMs, scoring a correlation of $0.1-0.4$ with the labels
Further, there is very little deviation between models with different initialisations.

\textbf{Discussion: }
In contrast to the Experiment on NNs, LMs are better able to learn from CTEs generated by \textsf{\small DaC} than by \textsf{\small MCTO}. This calls into question the conclusion that \textsf{\small MCTO} is the most effective generation algorithm. However, the generally bad performance of the LMs makes it relatively weak evidence about the effectiveness of methods. 

Overall the LM achieves a significantly lower performance on both tasks for all algorithms. Likely, the flexibility of the Neural Network enables it to learn more complex functions that explain the rewards. This also indicates that the Neural Network learned more complex knowledge about the reward function which is not straightforwardly learned by a LM. Furthermore, LMs do not benefit from cross-task learning, while the NN had a body that was shared between the contrastive and single tasks.

\end{document}